\documentclass[submission]{sacjsub}

\usepackage{wrapfig}
\usepackage{subfig}
\usepackage{amsfonts}
\usepackage{amsmath}
\usepackage{comment}
\usepackage{footmisc}
\usepackage{array}
\usepackage{tikz}
\usepackage{tablefootnote}

\begin{document}

\SACJtitle{A survey of benchmarking frameworks for reinforcement learning}

\SACJauthor[r]{Belinda Stapelberg}{belinda.stapelberg@up.ac.za}{corresponding}
\SACJauthor[r]{Katherine M. Malan}{malankm@unisa.ac.za}{}

%

\SACJaddress[r]{Department of Decision Sciences, University of South Africa, Pretoria, South Africa}

\SACJrunningheader{Stapelberg, B. and Malan, K.M.}{A survey of benchmarks for reinforcement learning algorithms}

\SACJcitationauthors{Belinda Stapelberg and Katherine M. Malan}

\SACJabstract{Reinforcement learning has recently experienced increased prominence in the machine learning community. There are many approaches to solving reinforcement learning problems with new techniques developed constantly. When solving problems using reinforcement learning, there are various difficult challenges to overcome. \par To ensure progress in the field, benchmarks are important for testing new algorithms and comparing with other approaches. The reproducibility of results for fair comparison is therefore vital in ensuring that improvements are accurately judged. This paper provides an overview of different contributions to reinforcement learning benchmarking and discusses how they can assist researchers to address the challenges facing reinforcement learning. The contributions discussed are the most used and recent in the literature. The paper discusses the contributions in terms of implementation, tasks and provided algorithm implementations with benchmarks. \par The survey aims to bring attention to the wide range of reinforcement learning benchmarking tasks available and to encourage research to take place in a standardised manner. Additionally, this survey acts as an overview for researchers not familiar with the different tasks that can be used to develop and test new reinforcement learning algorithms.

}

\SACJACMCategory{Computing methodologies}{Reinforcement learning}{h}
\SACJkeywords{reinforcement learning, benchmarking}

\SACJmaketitle

\section{Introduction}

Reinforcement learning (RL) is a subfield of machine learning, based on rewarding desired behaviours and/or punishing undesired ones of an agent interacting with its environment \cite{SuttonBarto_RLI_98}. The agent learns by taking sequential actions in its environment, observing the state of the environment and receiving a reward. The agent needs to learn a strategy, called a policy, to decide which action to take in any state. The goal of RL is to find the policy that maximises the long-term reward of the agent.

In recent years RL has experienced dramatic growth in research attention and interest due to promising results in areas including robotics control \cite{Lillicrap_et_al_15}, playing Atari 2600 \cite{Mnih_et_al_Atari_13, mnih2015human}, competitive video games \cite{Vinyals_et_al_Starcraft_17, Silva_Chaimowicz_Moba_17}, traffic light control \cite{Traffic_control} and more. In 2016, RL came into the general spotlight when Google DeepMind's AlphaGo \cite{Silver_et_al_GO_16} program defeated the Go world champion, Lee Sedol. Even more recently, Google DeepMind's AlphaStar AI program defeated professional StarCraft II players (considered to be one of the most challenging real-time strategy games) and OpenAI Five defeated professional Dota 2 players.

Progress in machine learning is driven by new algorithm development and the availability of high-quality data. In supervised and unsupervised machine learning fields, resources such as the UCI Machine Learning repository\footnote{\url{http://archive.ics.uci.edu/ml/index.php}}, the Penn Treebank \cite{marcus1993building}, the MNIST database of handwritten digits\footnote{\url{http://yann.lecun.com/exdb/mnist/}}, the ImageNet large scale visual recognition challenge \cite{Russakovsky2015}, and Pascal Visual Object Classes \cite{Everingham2010} are available. In contrast to the datasets used in supervised and unsupervised machine learning, progress in RL is instead driven by research on agent behaviour within challenging environments. Games have been used for decades to test and evaluate the performance of artificial intelligence systems. Many of the benchmarks that are available for RL are also based on games, such as the Arcade Learning Environment for Atari 2600 games \cite{ALE} but others involve tasks that simulate real-world situations, such as locomotion tasks in Garage (originally rllab) \cite{duan2016benchmarking}. These benchmarking tasks have been used extensively in research and significant progress has been made in using RL in ever more challenging domains.

Benchmarks and standardised environments are crucial in facilitating progress in RL. One advantage of the use of these benchmarking tasks is the reproducibility and comparison of algorithms to state-of-the-art RL methods. Progress in the field can only be sustained if existing work can be reproduced and accurately compared to judge improvements of new methods \cite{Henderson_et_al_18, machado2018revisiting}. The existence of standardised tasks can facilitate accurate benchmarking of RL performance.

This paper provides a survey of the most important and most recent contributions to benchmarking for RL. These are OpenAI Gym \cite{OpenAI_Gym}, the Arcade Learning Environment \cite{ALE}, a continuous control benchmark rllab \cite{duan2016benchmarking}, RoboCup Keepaway soccer \cite{stone2001keepaway} and Microsoft TextWorld \cite{TextWorld}. When solving RL problems, there are many challenges that need to be overcome, such as the fundamental trade-off problem between exploration and exploitation, partial observability of the environment, delayed rewards, enormous state spaces and so on. This paper discusses these challenges in terms of important RL benchmarking contributions and in what manner the benchmarks can be used to overcome or address these challenges.

The rest of the paper is organised as follows. Section~\ref{Sect:RL} introduces the key concepts and terminology of RL, and then discusses the approaches to solving RL problems and the challenges for RL. Section~\ref{Sect:Contributions} provides a survey on the contributions to RL benchmarking and Section~\ref{Sect:Summary} discusses the ways that the different contributions to RL benchmarking deal with or contribute to the challenges for RL. A conclusion follows in Section~\ref{Sect:Conclusion}.

\section{Reinforcement learning}\label{Sect:RL}

\begin{figure}
\begin{center}
\tikzstyle{block} = [rectangle, draw,
     text width=8em, text centered, rounded corners, minimum height=4em]

     \tikzstyle{line} = [draw, -latex]

     \begin{tikzpicture}[node distance = 6em, auto, thick]
     \node [block] (Agent) {Agent};
     \node [block, below of=Agent] (Environment) {Environment};

     \path [line] (Agent.0) --++ (4em,0em) |- node [near start]{Action
$a_t$} (Environment.0);
     \path [line] (Environment.190) --++ (-6em,0em) |- node [near start]
{New state  $s_{t+1}$} (Agent.170);
     \path [line] (Environment.170) --++ (-4.25em,0em) |- node [near
start, right] {Reward $r_{t+1}$} (Agent.190);

     \end{tikzpicture}
        \caption{Illustration of an RL system.}
     \label{RL_System}
     \end{center}
     \end{figure}
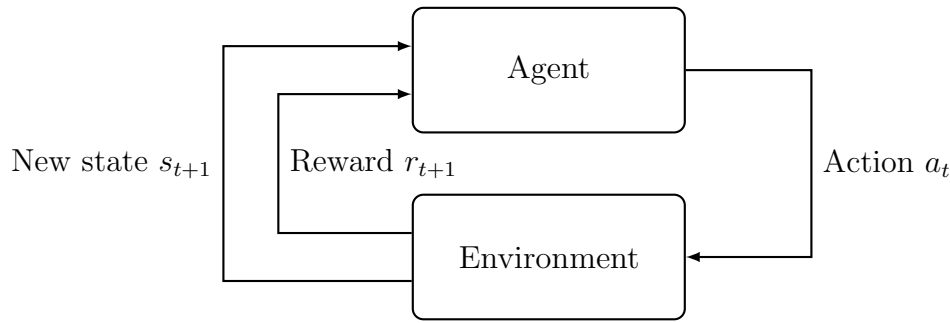

RL focuses on training an agent by using a trial-and-error approach. Figure~\ref{RL_System} illustrates the workings of an RL system. The agent evaluates a current situation (state), takes an action, and receives feedback (reward) from the environment after each act. The agent is rewarded with either positive feedback (when taking a ``good" action) or negative feedback as punishment for taking a ``bad" action. An RL agent learns how to act best through many attempts and failures. Through this type of trial-and-error learning, the agent's goal is to receive the best so-called long-term reward. The agent gets short-term rewards that together lead to the cumulative, long-term reward. The key goal of RL is to define the best sequence of actions that allow the agent to solve a problem while maximizing its cumulative long-term reward. That set of optimal actions is learned through the interaction of the agent with its environment and observation of rewards in every state.

This section provides the key concepts and terminology of RL used throughout this paper. The challenges of RL are also discussed.

\subsection{Concepts and terminology}

The core idea behind RL is to learn from the environment through interactions and feedback, and find an optimal strategy for solving the problem. The agent takes actions in its environment based on a (possibly partial) observation of the state of the environment and the environment provides a reward for the actions, which is usually a scalar value. The set of all valid actions is referred to as the action space, which can be either discrete (as in Atari and Go) or continuous (controlling a robot in the physical world). The goal of the agent is to maximise its long-term cumulative reward.

\subsubsection{Policy}

A policy of an agent is the control strategy used to make decisions, and is a mapping from states to actions. A policy can be deterministic or stochastic and is denoted by $\pi$. A deterministic policy maps states to actions without uncertainty while a stochastic policy is a probability distribution over actions for a given state. Therefore, when an agent follows a deterministic policy it will always take the same action for a given state, whereas a stochastic policy may take different actions in the same state. The immediate advantage of a stochastic policy is that an agent is not doomed to repeat a looped sequence of non-advancing actions.

\subsubsection{On-policy and off-policy learning}

There are two types of policy learning methods. On-policy learning is when the agent ``learns on the job'', i.e. it evaluates or improves the policy that is used to make the decisions directly. Off-policy learning is when the agent learns one policy, called the target policy, while following another policy, called the behaviour policy, which generates behaviour. The off-policy learning method is comparable to humans learning a task by observing others performing the task. 

\subsubsection{Value functions}

Having a value for a state (or state-action pair) is often useful in guiding the agent towards the optimal policy. The value under policy $\pi$ is the expected return if the agent starts in a specific state or state-action pair, and then follows the policy thereafter. So the state-value function $v_{\pi}$ is a mapping from states to real numbers and represents the long-term reward obtained by starting from a particular state and executing policy $\pi$. The action-value function $q_{\pi}$ is a mapping from state-action pairs to real numbers. The action-value $q_{\pi}(s, a)$ of state $s$ and action $a$ (where $a$ is an arbitrary action and not necessarily in line with the policy) is the expected return from starting in state $s$, taking action $a$ and then following policy $\pi$. The optimal value function $v_*$ gives the expected return starting in a state and then following the optimal policy $\pi_*$. The optimal action-value function $q_*$ is the expected return starting in some state, taking an arbitrary action and then following the optimal policy $\pi_*$.

These state-value and action-value functions all obey so-called Bellman equations, where the idea is that the value of the agent's starting point is the reward that is expected to be obtained from being there, plus the value of wherever the agent lands next. These Bellman equations are used in most RL approaches where the Bellman-backup is used, i.e. for a state or state-action pair the Bellman-backup is the (immediate) reward plus the next value.

\subsubsection{Function approximators}

In many RL problems the state space can be extremely large. Traditional solution methods where value functions are represented as arrays or tables mapping all states to values are therefore very difficult \cite{SuttonBarto_RLI_98}. One approach to this shortcoming is to use features to generalise an estimation of the value of states with similar features. Methods that compute these approximations are called function approximators. There are many techniques used for implementing function approximators including linear combinations of features, neural networks, decision trees, nearest neighbours, etc.

\subsubsection{Monte Carlo methods}

Monte Carlo methods are a class of learning methods where value functions are learned \cite{SuttonBarto_RLI_98}. The value of a state, $s_i$, is estimated by running many trials starting from $s_i$ and then averaging the total rewards received on those trials.

\subsubsection{Temporal difference algorithms}

Temporal difference (TD) learning algorithms are a class of learning methods that are based on the idea of comparing temporally successive predictions  \cite{SuttonBarto_RLI_98}. These methods are a fundamental idea in RL and use a combination of Monte Carlo learning and dynamic programming \cite{SuttonBarto_RLI_98}. TD methods learn value functions directly from experience by using the so-called TD error and bootstrapping (not waiting for a final outcome).

\subsubsection{Markov decisions processes}

The standard formalism for RL settings is called a Markov decision process (MDP). MDPs are used to define the interaction between an agent and its environment in terms of states, actions, and rewards. For an RL problem to be an MDP, it has to satisfy the Markov property: ``The future is independent of the past given the present''. This means that once the current state is known, then the history encountered so far can be discarded and that state completely characterises all the information needed as it captures all the relevant information from the history. Mathematically, an MDP is a tuple: $\langle \mathcal{S}, \mathcal{A}, \mathcal{R}, \mathcal{P}, \gamma \rangle$, where $\mathcal{S}$ is a (finite) set of states, $\mathcal{A}$ is a (finite) set of actions, $\mathcal{R}: \mathcal{S}\times \mathcal{A} \times \mathcal{S} \rightarrow \mathbb{R}$ is the reward function, $\mathcal{P}$ is a state transition probability matrix and $\gamma \in [0,1]$ is a discount factor included to control the reward.

\subsubsection{Model-free and model-based reinforcement learning approaches}\label{Subsect:Model_free_vs_based}

There are different aspects of RL systems that can be learnt. These include learning policies (either deterministic or stochastic), learning action-value functions (so-called Q-functions or Q-learning), learning state-value functions, and/or learning a model of the environment. A model of the environment is a function that predicts state transitions and rewards, and is an optional element of an RL system. If a model is available, i.e. if all the elements of the MDP are known, particularly the transition probabilities and the reward function, then a solution can be computed using classic techniques before executing any action in the environment. This is known as planning: computing the solution to a decision-making problem before executing an actual decision. 

When an agent does not know all the elements of the MDP, then the agent does not know how the environment will change in response to its actions or what its immediate reward will be. In this situation the agent will have to try out different actions, observe what happens and in some way find a good policy from doing this. One approach to solve a problem without a complete model is for the agent to learn a model of how the environment works from its observations and then plan a solution using that model. Methods that use the framework of models and planning are referred to as model-based methods. 

Another way of solving RL problems without a complete model of the environment is to learn through trial-and-error. Methods that do not have or learn a model of the environment and do not use planning are called model-free methods. The two main approaches to represent and train agents with model-free RL are policy optimisation and Q-learning. In policy optimisation methods (or policy-iteration methods) the agent learns the policy function directly. Examples include policy gradient methods, asynchronous advantage actor-critic (A3C) \cite{mnih2016asynchronous}, trust region policy optimization (TRPO) \cite{Schulman_etal_2015} and proximal policy optimization (PPO) \cite{PPO_17}. Q-Learning methods include deep Q-networks (DQN) \cite{Mnih_et_al_Atari_13}, C51 algorithm \cite{bellemare2017distributional} and Hindsight Experience Replay (HER) \cite{andrychowicz2017hindsight}. Hybrid methods combining the strengths of Q-learning and policy gradients exist as well, such as deep deterministic policy gradients (DDPG) \cite{Lillicrap_et_al_15}, soft actor-critic algorithm (SAC) \cite{haarnoja2018soft} and twin delayed deep deterministic policy gradients (TD3) \cite{fujimoto2018addressing}.

In the current literature, the most used approaches incorporates a mixture of model-based and model-free methods, such as Dyna and Monte Carlo tree search (MCTS) \cite{SuttonBarto_RLI_98}, and temporal difference search \cite{Silver2012}.

\subsection{Challenges for reinforcement learning}\label{Subsect:Challenges}

This section discusses some of the challenges faced by RL. These challenges will be discussed in terms of how they are addressed by different contributions in Section \ref{Sect:Summary}.

\subsubsection{Partially observable environment}

How the agent observes the environment can have a significant impact on the difficulty of the problem. In most real-world environments the agent does not have a complete or perfect perception of the state of its environment due to incomplete information provided by its sensors, the sensors being noisy or some of the state being hidden. However, for learning methods that are based on MDPs, the complete state of the environment should be known. To address the problem of partial observability of the environment, the MDP framework is extended to the partially observable Markov decision process (POMDP) model.

\subsubsection{Delayed or sparse rewards} 

In an RL problem, an agent's actions determine its immediate reward as well as the next state of the environment. Therefore, an agent has to take both these factors into account when deciding which action to take in any state. Since the goal is to learn which actions to take that will give the most reward in the long-run, it can become challenging when there is little or no immediate reward. The agent will consequently have to learn from delayed reinforcement, where it may take many actions with insignificant rewards to reach a future state with full reward feedback. The agent must therefore be able to learn which actions will result in an optimal reward, which it might only receive far into the future. 

In line with the challenge of delayed or sparse rewards is the problem of long-term credit assignment \cite{minsky1961steps}: how must credit for success be distributed among the sequence of decisions that have been made to produce the outcome?

\subsubsection{Unspecified or multi-objective reward functions}

Many tasks (especially real-world problems) have multiple objectives. The goal of RL is to optimise a reward function, which is commonly framed as a global reward function, but tasks with more than one objective could require optimisation of different reward functions. In addition, when an agent is training to optimise some objective, other objectives could be discovered which might have to be maintained or improved upon. Work on multi-objective RL (MORL) has received increased interest, but research is still primarily devoted to single-objective RL.

\subsubsection{Size of the state and action spaces}

Large state and action spaces can result in enormous policy spaces in RL problems. Both state and action spaces can be continuous and therefore infinite. However, even discrete states and actions can lead to infeasible enumeration of policy/state-value space. In RL problems for which state and/or action spaces are small enough, so-called tabular solutions methods can be used, where value functions can be represented as arrays or tables and exact solutions are often possible. For RL problems with state and/or action spaces that are too large, the goal is to instead find good approximate solutions with the limited computational resources available and to avoid the curse of dimensionality \cite{Bellman_1957}.

\subsubsection{The trade-off between exploration and exploitation}

One of the most important and fundamental overarching challenges in RL is the trade-off between exploration and exploitation. Since the goal is to obtain as much reward as possible, an agent has to learn to take actions that were previously most effective in producing a reward. However, to discover these desirable actions, the agent has to try actions that were not tried before. It has to exploit the knowledge of actions that were already taken, but also explore new actions that could potentially be better selections in the future. The agent may have to sacrifice short-term gains to achieve the best long-term reward. Therefore, both exploration and exploitation are fundamental in the learning process, and exclusive use of either will result in failure of the task at hand. There are many exploration strategies \cite{SuttonBarto_RLI_98}, but a key issue is the scalability to more complex or larger problems. The exploration vs. exploitation challenge is affected by many of the other challenges that are discussed in this section, such as delayed or sparse rewards, and the size of the state or action spaces.

\subsubsection{Representation learning}

Representation (or feature) learning involves automatically extracting features or understanding the representation of raw input data to perform tasks such as classification or prediction. It is fundamental not just to RL, but to machine learning and AI in general, even with a conference dedicated to it: International Conference on Learning Representations (ICLR). 

One of the clearest challenges that representation learning tries to solve in an RL context is to effectively reduce the impact of the curse of dimensionality, which results from very large state and/or action spaces. Ideally an effective representation learning scheme will be able to extract the most important information from the problem input in a compressed form.

\subsubsection{Transfer learning}

Transfer learning \cite{Pan_Yang_Transfer_2010, weiss2016survey} uses the notion that, as in human learning, knowledge gained from a previous task can improve the learning in a new (related) task through the transfer of knowledge that has already been learned. The field of transfer learning has recently been experiencing growth in RL \cite{taylor2009transfer} to accelerate learning and mitigate issues regarding scalability.

\subsubsection{Model learning}

Model-based RL methods (Section~\ref{Subsect:Model_free_vs_based}) are important in problems where the agent's interactions with the environment are expensive. These methods are also significant in the trade-off between exploration and exploitation, since planning impacts the need for exploration. Model learning can reduce the interactions with the environment, something which can be limited in practice, but introduces additional complexities and the possibility of model errors. Another challenge related to model learning is the problem of planning using an imperfect model, which is also a difficult challenge that has not received much attention in the literature.

\subsubsection{Off-policy learning}

Off-policy learning methods (e.g. Q-learning) scale well in comparison to other methods and the  algorithms can (in principle) learn from data without interacting with the environment. An agent is trained using data collected by other agents (off-policy data) and data it collects itself to learn generalisable skills. 

Disadvantages of off-policy learning methods include greater variance and slow convergence, but are more powerful and general than on-policy learning methods \cite{SuttonBarto_RLI_98}.  Advantages of using off-policy learning is the use of a variety of exploration strategies, and learning from training data that are generated by unrelated controllers, which includes manual human control and previously collected data.

\subsubsection{Reinforcement learning in real-world settings}

The use of RL in real-world scenarios has been gaining attention due to the success of RL in artificial domains. In real-world settings, more challenges become apparent for RL. Dulac-Arnold \textit{et al.} \cite{Real_world_RL_challenges} provide a list of nine challenges for RL in the real-world, many of which have been mentioned in this section already. Further challenges not discussed here include \textit{safety constraints}, \textit{policy explainability} and \textit{real-time inference}. Many of these challenges have been studied extensively in isolation, but there is a need for research on algorithms (both in artificial domains and real-world settings) that addresses more than one or all of these challenges together, since many of the challenges are present in the same problem.

\subsubsection{A standard methodology for benchmarking}

A diverse range of methodologies is currently common in the literature, which brings into question the validity of direct comparisons between different approaches. A standard methodology for benchmarking is necessary for the research community to compare results in a valid way and accelerate advancement in a rigorous scientific manner.


\section{Contributions to reinforcement learning benchmarking}\label{Sect:Contributions}

This section discusses some important reinforcement learning benchmarks currently in use. The list of contributions is by no means exhaustive, but includes the ones that are most in use currently in the RL research community.


\subsection{OpenAI Gym}\label{Sect:OpenAI_Gym}

Released publicly in April 2016, OpenAI's Gym \cite{OpenAI_Gym} is a toolkit for developing and comparing reinforcement learning algorithms. It includes a collection of benchmark problems which is continuing to grow as well as a website where researchers can share their results and compare algorithm performance. It provides a tool to standardise reporting of environments in research publications to facilitate the reproducibility of published research. OpenAI Gym has become very popular since its release, with \cite{OpenAI_Gym} having over 1300 citations on Google Scholar to date.

\subsubsection{Implementation}

The OpenAI Gym library is a collection of test problems (environments) with a common interface and makes no assumptions about the structure of an agent. OpenAI Gym currently supports Linux and OS X running Python 2.7 or 3.5 -- 3.7. Windows support is currently experimental, with limited support for some problem environments. OpenAI Gym is compatible with any numerical computation library, such as TensorFlow or Theano. To get started with OpenAI Gym, visit the documentation site\footnote{\url{https://gym.openai.com}\label{Gym_main}} or the actively maintained GitHub repository\footnote{\url{https://github.com/openai/gym}}.

\subsubsection{Benchmark tasks}

The environments available in the library are diverse, ranging from easy to difficult and include a variety of data. A brief overview of the different environments is provided here with the full list and descriptions of environments available on the main site\footref{Gym_main}.

\begin{figure}[h!]
\subfloat[A screenshot of the classic control task Cart-Pole, with the objective to keep the pole balanced by moving the cart.\label{Cart_pole}]
{\includegraphics[width=0.3\textwidth]{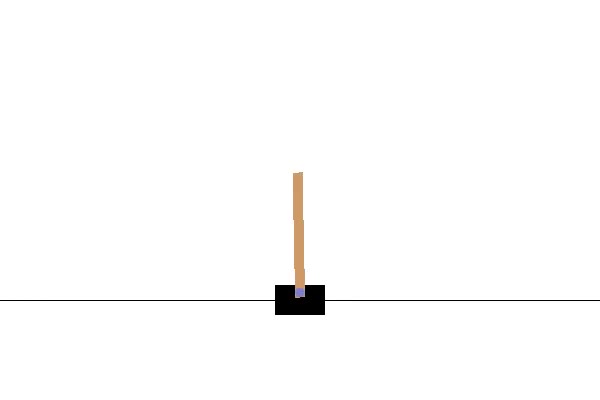}}
     \hfill
\subfloat[A screenshot of the Atari 2600 game Breakout.\label{Atari_Breakout}]
{\includegraphics[width=0.3\textwidth]{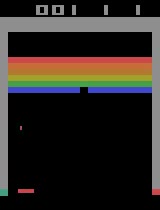}}
     \hfill
\subfloat[A screenshot of the MuJoCo simulator, where a four-legged 3D robot has to learn to walk.\label{big_ant}]
{\includegraphics[width=0.3\textwidth]{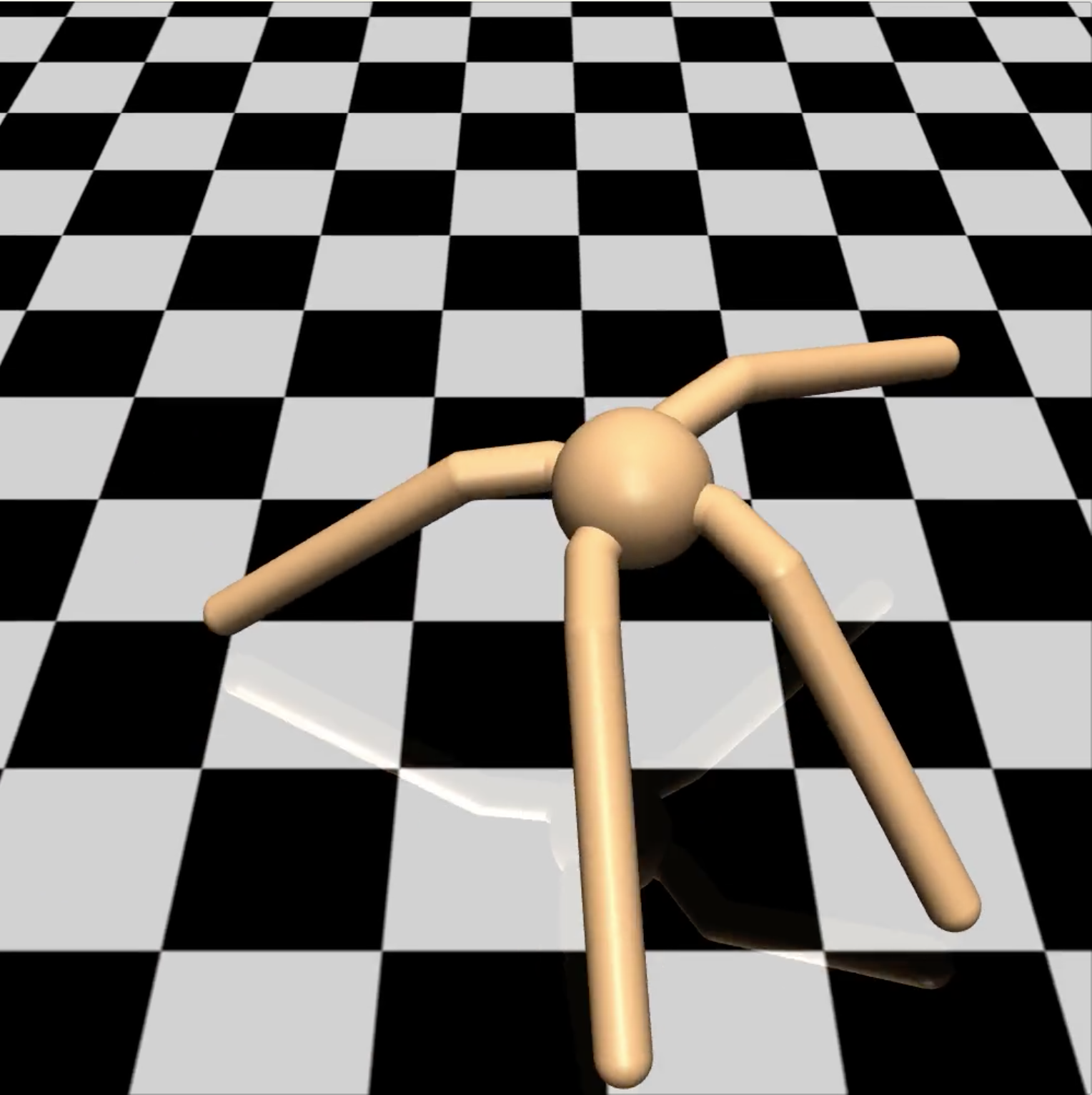}}
     \hfill         
\subfloat[A screenshot of the Box2D simulator, where a bipedal robot has to learn to walk.\label{box2D_bipedal}]
{\includegraphics[width=0.3\textwidth]{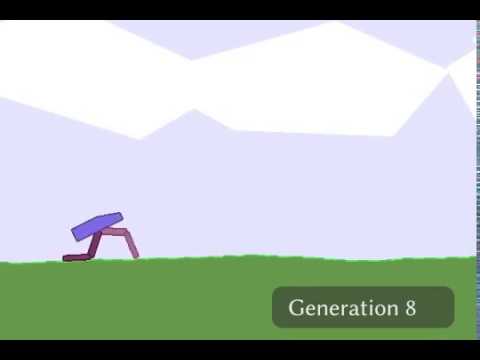}}
     \hfill  
\subfloat[A screenshot of the 3D humanoid robot learning to walk as fast as possible in the Roboschool simulator.\label{roboschool_atlas}]
{\includegraphics[width=0.3\textwidth]{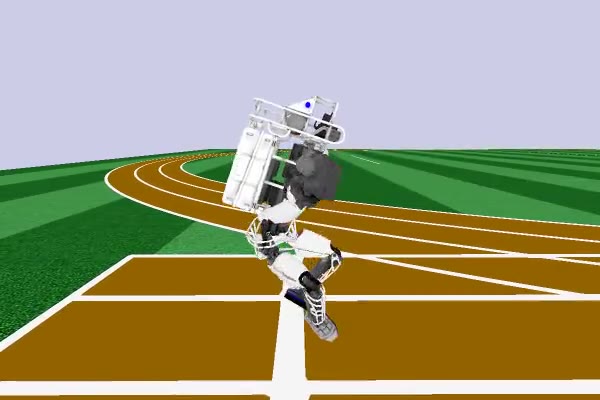}}
     \hfill  
\subfloat[A screenshot of the ShadowHand robot manipulating a block.\label{hand-block}]
{\includegraphics[width=0.3\textwidth]{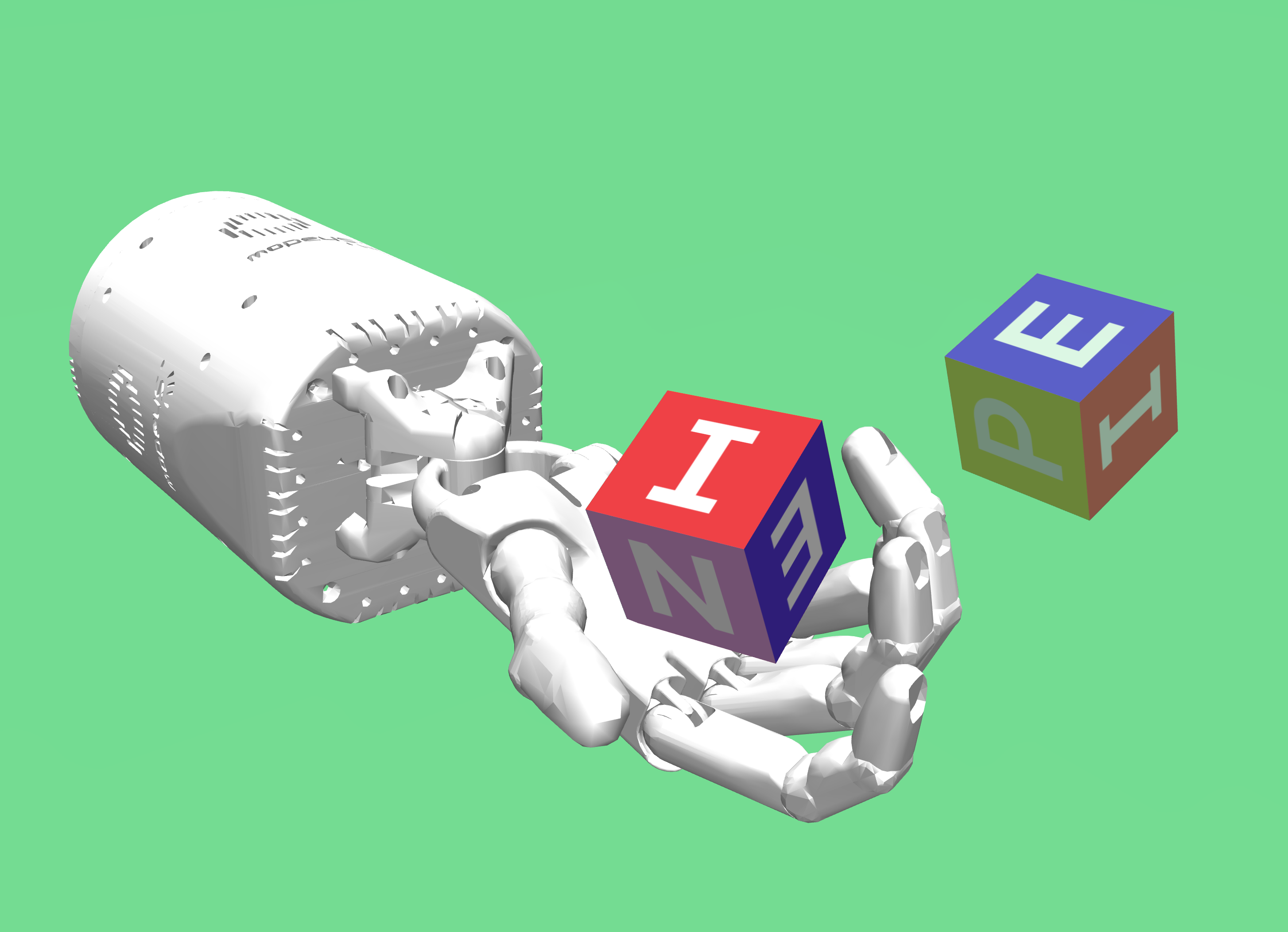}}
     \hfill  
     \caption{Some examples of the environments used in OpenAI Gym.}
     \label{OpenAI_Gym_Pictures}   
      
\end{figure}

\textbf{Classic control and toy text}: These small-scale problems are a good starting point for researchers not familiar with the field. The classic control problems include balancing a pole on a moving cart (Figure~\ref{Cart_pole}), driving a car up a steep hill, swinging a pendulum and more. The toy text problems include finding a safe path across a grid of ice and water tiles, playing Roulette, Blackjack and more.

\textbf{Algorithmic}: The objective here is for the agent to learn algorithms such as adding multi-digit numbers and reversing sequences, purely from examples. The difficulty of the tasks can be varied by changing the sequence length.

\textbf{Atari 2600}: The Arcade Learning Environment (ALE) \cite{ALE} has been integrated into OpenAI Gym in easy-to-install form, where classic Atari 2600 games (see Figure~\ref{Atari_Breakout} for an example) can be used for developing agents (see Section~\ref{Sect:ALE} for a detailed discussion). For each game there are two versions: a version which takes the RAM as input and a version which takes the observable screen as the input. 


\textbf{MuJoCo}: These robot simulation tasks use the MuJoCo proprietary software physics engine \cite{MuJoCo_citation}, but free trial and postgraduate student licences are available. The problems include 3D robot walking or standing up tasks, 2D robots running, hopping, swimming or walking (see Figure~\ref{big_ant} for an example), balancing two poles vertically on top of each other on a moving cart, and repositioning the end of a two-link robotic arm to a given spot.

\textbf{Box2D}: These are continuous control tasks in the Box2D simulator, which is a free open source 2-dimensional physics simulator engine. Problems include training a bipedal robot (Figure~\ref{box2D_bipedal}) to walk (even on rough terrain), racing a car around a track and navigating a lunar lander to its landing pad.

\textbf{Roboschool}: Most of these problems are the same as in MuJoCo, but use the open-source software physics engine, Bullet. Additional tasks include teaching a 3D humanoid robot to walk as fast as possible (see Figure~\ref{roboschool_atlas}) as well as a continuous control version of Atari Pong.

\textbf{Robotics}: Released in 2018, these environments are used to train models which work on physical robots. It includes four environments using the Fetch\footnote{\url{https://fetchrobotics.com/}} research platform and four environments using the ShadowHand\footnote{\url{https://www.shadowrobot.com/products/dexterous-hand/}} robot. These manipulation tasks are significantly more difficult than the MuJoCo continuous control environments. The tasks for the Fetch robot are to move the end-effector to a desired goal position, hitting a puck across a long table such that it slides and comes to rest on the desired goal, moving a box by pushing it until it reaches a desired goal position, and picking up a box from a table using its gripper and moving it to a desired goal above the table. The tasks for the ShadowHand are reaching with its thumb and a selected finger until they meet at a desired goal position above the palm, manipulating a block (see Figure~\ref{hand-block}), an egg, and a pen, until the object achieves a desired goal position and rotation.

Alongside these new robotics environments, OpenAI also released code for Hindsight Experience Replay (HER), a reinforcement learning algorithm that can learn from failure. Their results show that HER can learn successful policies on most of the new robotics problems from only sparse rewards. A set of requests for research has also been released\footnote{\url{https://openai.com/blog/ingredients-for-robotics-research/}\label{Request_Robotics}} in order to encourage and facilitate research in this area, with a few ideas of ways to improve HER specifically.

\subsection{The Arcade Learning Environment}\label{Sect:ALE}

The Atari 2600 gaming console was released in September 1977, with over 565 games developed for it over many different genres. The games are considerably simpler than modern era video games. However, the Atari 2600 games are still challenging and provide interesting tasks for human players.

The Arcade Learning Environment (ALE) \cite{ALE} is an object-oriented software framework allowing researchers to develop AI agents for the original Atari 2600 games. It is a platform to empirically assess and evaluate AI agents designed for general competency. ALE allows interfacing through the Atari 2600 emulator Stella and enables the separation of designing an AI agent and the details of emulation. There are currently over 50 game environments supported in the ALE.

The ALE has received a lot of attention since its release in 2013 (over 1200 citations on Google Scholar to date), perhaps the most note-worthy being the success of Deep Q-networks (DQN), which was the first algorithm to achieve human-level control performance in many of the Atari 2600 games \cite{mnih2015human}.

\subsubsection{Implementation}

The Stella emulator interfaces with the Atari 2600 games by receiving joystick movements and sending screen and/or RAM information to the user. For the reinforcement learning context, ALE has a game-handling layer to provide the accumulated score and a signal for whether the game has ended. The default observation of a single game screen or frame is made up of a two-dimensional array of 7-bit pixels, 160 pixels wide by 210 pixels high. The joystick controller defines 18 discrete actions, which makes up the action space of the problem. Only some actions are needed to play a game and the game-handling layer also provides the minimum set of actions needed to play any particular game. The simulator generates 60 frames per second in real-time and up to 6000 frames per second at full speed. The reward the agent receives depends on each game, but is generally the score difference between frames. A game episode starts when the first frame is shown and ends when the goal of the game has been achieved or after a predefined number of frames. The ALE therefore offers access to a variety of games through one common interface.

The ALE also has the functionality of saving and restoring the current state of the emulator. This functionality allows the investigation of topics including planning and model-based reinforcement learning.

ALE is free, open-source software\footnote{\url{http://arcadelearningenvironment.org}}, including the source code for the agents used in associated research studies \cite{ALE}. ALE is written in C++, but there are many interfaces available that allow the interaction with ALE in other programming languages, with detail provided in \cite{ALE}.

Due to the increase in popularity and importance in the AI literature, another paper was published in 2018 by some of the original proposers of the ALE \cite{machado2018revisiting}, providing a broad overview of how the ALE is used by researchers, highlighting overlooked issues and discussing propositions for maximising the future use of the testbed. Concerns are raised at how agents are evaluated in the ALE and new benchmark results are provided. 

In addition, a new version of the ALE was introduced in 2018 \cite{machado2018revisiting}, which supports multiple game modes and includes so called \textit{sticky actions}, providing some form of stochasticity to the controller. When sticky actions are used, there is a possibility that the action requested by the agent is not executed, but instead the agent's previous action is used, emulating a sticky controller. The probability that an action will be sticky can be specified using a pre-set control parameter. The original ALE is fully deterministic and consequently it is possible for an agent to memorise a good action sequence, instead of learning how to make good decisions. Introducing sticky actions therefore increases the robustness of the policy that the agent has to learn.

Originally the ALE only allowed agents to play games in their default mode and difficulty. In the latest version of the ALE  \cite{machado2018revisiting} it is possible to select among different game modes and difficulty levels for single player games, where each mode-difficulty pair is referred to as a \textit{flavour}. Changes in the mode and difficulty of the games can impact game dynamics and introduce new actions.

\subsubsection{Published benchmark results}

Bellemare \textit{et al.} \cite{ALE} provide performance results on the ALE tasks using an augmented version of the SARSA($\lambda$) \cite{SuttonBarto_RLI_98} algorithm, where linear function approximation is used. For comparison, the performance results of a non-expert human player and three baseline agents (Random, Const and Perturb) are also provided. A set of games is used for training and parameter tuning, and another set for testing. The ALE can also be used to study planning techniques. Benchmark results for two traditional search methods (Breadth-first search and UCT: Upper Confidence Bounds Applied to Trees) are provided, as well as the performance results of the best learning agent and the best baseline policy.

Machado \textit{et al.} \cite{machado2018revisiting} provide benchmark results for 60 Atari 2600 games with sticky actions for DQN and SARSA($\lambda$) + Blob-PROST \cite{liang2016state} (an algorithm that includes a feature representation which enables SARSA($\lambda$) to achieve performance that is comparable to that of DQN).


\subsection{Continuous control: rllab}\label{Sect:Cont_Control}

The Arcade Learning Environment (Section~\ref{Sect:ALE}) is a popular benchmark to evaluate algorithms which are designed for tasks with discrete actions. Duan \textit{et al.} \cite{duan2016benchmarking} present a benchmark of 31 continuous control tasks, ranging in difficulty, and also implement a range of RL algorithms on the tasks. 

The benchmark as well as the implementations of the algorithms are available at the rllab GitHub repository\footnote{\url{https://github.com/rll/rllab}}, however this repository is no longer under development but is currently actively maintained at the \textbf{garage} GitHub repository\footnote{\url{https://github.com/rlworkgroup/garage}}, which includes many improvements. The documentation\footnote{\url{https://garage.readthedocs.io/en/latest/}} for garage is a work in progress and the available documentation is currently limited. Both rllab and garage are fully compatible with OpenAI Gym and only support Python 3.5 and higher. 

Other RL benchmarks for continuous control have also been proposed, but many are not in use anymore. Duan \textit{et al.} \cite{duan2016benchmarking} provide a comprehensive list of benchmarks containing low-dimensional tasks as well as a wide range of tasks with high-dimensional continuous state and action spaces. They also discuss previously proposed benchmarks for high-dimensional control tasks do not include such a variety of tasks as in rllab. Where relevant, we mention some of these benchmarks in the next section that have additional interesting tasks.

\subsubsection{Benchmark tasks}\label{rllab:Tasks}

There are four categories for the rllab continuous control tasks: basic, locomotion, partially observable and hierarchical tasks.

\textbf{Basic tasks:} These five tasks are widely analysed in the reinforcement learning and control literature. Some of these tasks can also be found in the ``Classic control" section of OpenAI Gym (Section~\ref{Sect:OpenAI_Gym}). The tasks are cart-pole balancing, cart-pole swing up, mountain car, acrobot swing up and double inverted pendulum balancing (which can be found in OpenAI Gym Roboschool). A related benchmark involving a 20 link pole balancing task is proposed as part of the Tdlearn package \cite{JMLR:v15:dann14a}. 

\textbf{Locomotion tasks:} Six locomotion tasks of varying dynamics and difficulty are implemented with the goal to move forward as quickly as possible. These tasks are challenging due to high degrees of freedom as well as the need for a lot of exploration, since getting stuck at a local optima (such as staying at the origin or diving forward slowly) can happen easily when the agent acts greedily. These tasks are: Swimmer, Hopper, Walker, Half-Cheetah, Ant, Simple Humanoid and Full Humanoid.

Other environments with related locomotion tasks include dotRL \cite{papis2013dotrl} with a variable segment octopus arm \cite{woolley_stanley_octupus_arm_2010}, PyBrain \cite{schaul2010pybrain}, and SkyAI \cite{yamaguchi2010skyai} with humanoid robot tasks like jumping, crawling and turning.

\textbf{Partially observable tasks:} Realistic agents often do not have access to perfect state information due to limitations in sensory input. To address this, three variations of partially observable tasks are implemented for each of the five basic tasks mentioned above. This leads to 15 additional tasks. The three variations are \textit{limited sensors} (only positional information is provided, no velocity), \textit{noisy observations and delayed actions} (Gaussian noise is added to simulate sensor noise, and a time delay is added between taking an action and an action being executed) and \textit{system identification} (the underlying physical model parameters vary across different episodes). These variations are not currently available in OpenAI Gym.

\textbf{Hierarchical tasks:} In many real-world situations higher level decisions can reuse lower level skills, for example a robot learning to navigate a maze can reuse learned locomotion skills. Here tasks are proposed where low-level motor controls and high-level decisions are needed, which operate on different time scales and a natural hierarchy exists in order to learn the task most efficiently. The tasks are as follows. \textit{Locomotion and food collection}: where the swimmer or the ant robot operates in a finite region and the goal is to collect food and avoid bombs. \textit{Locomotion and maze}: the swimmer or the ant robot has the objective to reach a specific goal location in a fixed maze environment. These tasks are not currently available in OpenAI Gym.

\subsubsection{Published benchmark results}

Duan \textit{et al.} \cite{duan2016benchmarking} provide performance results on the rllab tasks. The algorithms implemented are mainly gradient-based policy search methods, but two gradient-free methods are included for comparison. Almost all of the algorithms are batch algorithms and one algorithm is an online algorithm. The batch algorithms are REINFORCE \cite{Williams92_REINFORCE}, truncated natural policy gradient (TNPG) \cite{duan2016benchmarking}, reward-weighted regression (RWR) \cite{Peters_Schaal_07}, relative entropy policy search (REPS) \cite{AAAI101851}, trust region policy optimization (TRPO) \cite{Schulman_etal_2015}, cross entropy method (CEM) \cite{Rubinstein1999} and covariance matrix adaptation evolution strategy (CMA-ES) \cite{Hansen_Ostermeier_2001}. The online algorithm used is deep deterministic policy gradient (DDPG) \cite{Lillicrap_et_al_15}. Direct applications of the batch-based algorithms to recurrent policies are implemented with minor modifications.

Of the implemented algorithms, TNPG, TRPO and DDPG were effective in training deep neural network policies. However, all algorithms performed poorly on the hierarchical tasks, which suggest that new algorithms should be developed for automatic discovery and exploitation of the tasks' hierarchical structure.

Recently a new class of reinforcement learning algorithms called proximal policy optimisation (PPO) \cite{PPO_17} was released by OpenAI. PPO's performance is comparable or better than state-of-the-art approaches to solving 3D locomotion, robotic tasks (similar to the tasks in the benchmark discussed above) and also Atari 2600, but it is simpler to implement and tune. OpenAI has adopted PPO as its go-to RL algorithm, since it strikes a balance between ease of implementation, sample complexity, and ease of tuning.


\subsection{RoboCup Keepaway Soccer}\label{Robocup}

RoboCup \cite{RoboCup_97} simulated soccer has been used as the basis for successful international competitions and research challenges since 1997. Keepaway is a subtask of RoboCup that was put forth as a testbed for machine learning in 2001 \cite{stone2001keepaway}. It has since been used for research on temporal difference reinforcement learning with function approximation \cite{stone2005reinforcement}, evolutionary learning \cite{pietro2002learning}, relational reinforcement learning \cite{walker2004relational}, behaviour transfer \cite{taylor2005behavior, didi2016multi, didi2016hybridizing, nitschke2017evolutionary, Didi:2018:robocup, Schwab:2018:transfer_robocup, sym11010025}, batch reinforcement learning \cite{riedmiller2009reinforcement} and hierarchical reinforcement learning \cite{bai2017efficient}.

\begin{wrapfigure}{L}{0.5\textwidth}
\centerline{\includegraphics[width=7cm]{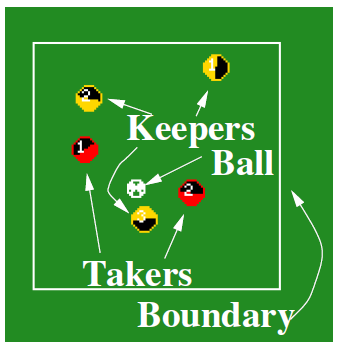}}
\caption{A screen shot from a 3v2 keepaway episode in a $20m \times 20m$ region from \cite{stone2001keepaway}.}
\label{Fig:Keepaway}
\end{wrapfigure}

In Keepaway, one team (the keepers) tries to maintain possession of the ball within a limited region, while the opposing team (the takers) attempts to gain possession \cite{stone2001keepaway}. The episode ends whenever the takers take possession of the ball or the ball leaves the region. The players are then reset for another episode with the keepers being given possession of the ball again. Task parameters include the size of the region, the number of keepers, and the number of takers. Figure~\ref{Fig:Keepaway} shows an example episode with 3 keepers and 2 takers (called 3v2) playing in a $20m \times 20m$ region \cite{stone2001keepaway}.

In 2005 Stone \textit{et al.} \cite{stone2005keepaway} elevated the Keepaway testbed to a benchmark problem for machine learning and provided infrastructure to easily implement the standardised task. 

An advantage of the Keepaway subtask is that it allows for direct comparison of different machine learning algorithms. It is also good for benchmarking machine learning since the task is simple enough to be solved successfully, but complex enough that straightforward solutions are not sufficient.

\subsubsection{Implementation}

A standardized Keepaway player framework is implemented in C++ and the source code is available for public use at an online repository\footnote{\label{note1}\url{http://www.cs.utexas.edu/~AustinVilla/sim/keepaway/}}. The repository provides implementation for all aspects of the Keepaway problem except the learning algorithm itself. It also contains a step-by-step tutorial of how to use the code, with the goal of allowing researchers who are not experts in the RoboCup simulated soccer domain to easily become familiar with the domain.

\subsubsection{Standardised task}

Robocup simulated soccer (and therefore also Keepaway) is a fully distributed, multiagent domain with both teammates and adversaries \cite{stone2000layered}. The environment is partially observable for each agent and the agents also have noisy sensors and actuators. Therefore, the agents do not perceive the world exactly as it is, nor can they affect the world exactly as intended. The perception and action cycles of the agent are asynchronous, therefore perceptual input does not trigger actions as is traditional in AI. Communication opportunities are limited, and the agents must make their decisions in real-time. These domain characteristics all result in simulated robotic soccer being a realistic and challenging domain \cite{stone2000layered}.

The size of the Keepaway region, the number of keepers, and the number of takers can easily be varied to change the task. Stone \textit{et al.} \cite{stone2005keepaway} provide a framework with a standard interface to the learner in terms of macro-actions, states, and rewards.

\subsubsection{Published benchmark results}

Stone \textit{et al.} \cite{stone2005keepaway} performed an empirical study for learning Keepaway by training the keepers using episodic SMDP SARSA($\lambda$) \cite{stone2005reinforcement, SuttonBarto_RLI_98}, with three different function approximators: CMAC function approximation \cite{albus1975new, Albus_1981}, Radial Basis Function (RBF) \cite{SuttonBarto_RLI_98} networks (a novel extension to CMACs \cite{stone2005keepaway}), and neural network function approximation. The RBF network performed comparably to the CMAC method. The Keepaway benchmark structure allows for these results to be quantitatively compared to other learning algorithms to test the relative benefits of different techniques.

\subsubsection{Half Field Offense: An extension to Keepaway}\label{HFO}

Half Field Offense (HFO) \cite{HFO_2007, hausknecht2016half} is an extension of Keepaway, which is played on half of the soccer field with more players on each team. The task was originally introduced in 2007 \cite{HFO_2007}, but no code was made publicly available. In 2016 \cite{hausknecht2016half} the HFO environment was released publicly (open-source)\footnote{\label{note3}\url{https://github.com/LARG/HFO}}, however this repository is not currently being maintained. 

Success in HFO means that the offensive players have to keep possession of the ball (the same as in Keepaway), learn to pass or dribble to get closer to the goal and shoot when possible. Agents can also play defence where they have to prevent goals from being scored. HFO also supports multi-agents which could be controlled manually or automatically.

In the same way as the Keepaway environment \cite{stone2005keepaway}, the HFO environment allows ease of use in developing and deploying agents in different game scenarios, with C++ and Python interfaces. The performance of three benchmark agents are compared in \cite{hausknecht2016half}, namely a random agent, a handcoded agent and a SARSA agent.

A similar platform to the Arcade Learning Environment (Section~\ref{Sect:ALE}), the HFO environment places less emphasis on generality (the main goal of the ALE) and more emphasis on cooperation and multiagent learning.


\subsection{Microsoft TextWorld}

Recently, researchers from the Microsoft Research Montreal Lab released an open source project called TextWorld \cite{TextWorld}, which attempts to train reinforcement learning agents using text-based games. 

In a time where AI agents are mastering complex multi-player games such as Dota 2 and StarCraft II, it might seem unusual to do research on text-based games. Text-based games can play a similar role to multi-player graphic environments which train agents to learn spatial and time-based planning, in advancing conversational skills such as affordance extraction (identifying which verbs are applicable to a given object), memory and planning, exploration etc. Another powerful motivation for the interest in text-based games is that language abstracts away complex physical processes, such as a robot trying not to fall over due to gravity. Text-based games require language understanding and successful play requires skills like long-term memory and planning, exploration (trial and error), common sense, and learning with these challenges. 

TextWorld is a sandbox environment which enables users to handcraft or automatically generate new games. These games are complex and interactive simulations where text is used to describe the game state and players enter text commands to progress though the game. Natural language is used to describe the state of the world, to accept actions from the player, and to report subsequent changes in the environment. The games are played through a command line terminal and are turn-based, i.e. the simulator describes the state of the game through text and then a player enters a text command to change its state in some desirable way.

\subsubsection{Implementation}

\begin{wrapfigure}{L}{0.575\textwidth}
\includegraphics[width = 10cm]{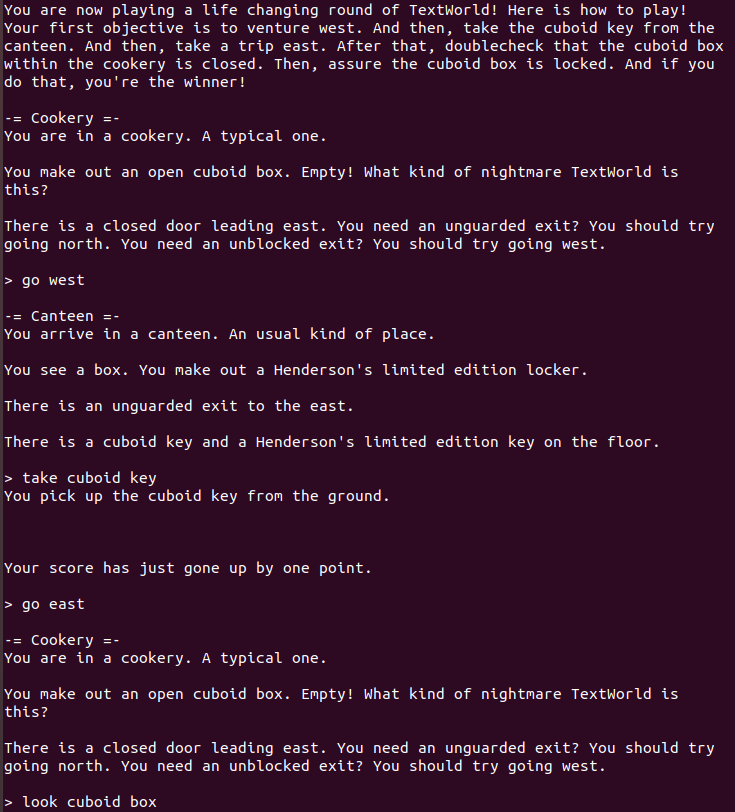}
\caption{An example game generated by TextWorld to illustrate the command structure of a game.}
\label{TextWorld:Command_Structure}
\end{wrapfigure}

In Figure~\ref{TextWorld:Command_Structure} an example game is shown in order to illustrate the command structure of a typical text-based game generated by TextWorld. 

TextWorld enables interactive play-through of text-based games and, unlike other text-based environments such as TextPlayer\footnote{\url{https://github.com/danielricks/textplayer}} and PyFiction\footnote{\url{https://github.com/MikulasZelinka/pyfiction}}, enables users to handcraft games or to construct games automatically. The TextWorld logic engine automatically builds game worlds, populates them with objects and obstacles, and generates quests that define a goal state and how to reach it \cite{TextWorld}. TextWorld requires Python 3 and currently only supports Linux and macOS systems. The code and documentation are available publicly\footnote{\label{Textworld_url}\url{http://aka.ms/textworld}} and the learning environment is described in full detail in Section 3 of \cite{TextWorld}, including descriptions of the two main components of the Python framework: a game generator and a game engine. To interact with TextWorld, the framework provides a simple application programming interface (API) which is inspired by OpenAI Gym.

In an RL context, TextWorld games can be seen as partially observable Markov decision processes. The environment state at any turn $t$ contains a complete description of the game state, but much of this is hidden from the agent. Once an agent has issued a command (of at least one word), the environment transitions to a next state with a certain probability. Since the interpreter in parser-based games can accept any sequence of characters (of any length), but only a fraction thereof is recognised, the resulting action space is very large. Therefore, two simplifying assumptions are made in \cite{TextWorld}: the commands are sequences of at most $L$ words taken from a fixed vocabulary $V$ and the commands have to follow a specific structure: a verb, a noun phrase and an adverb phrase. The action space of the agent is therefore the set of all permissible commands from the fixed vocabulary $V$ followed by a certain special token (``enter'') that signifies the end of the command.

The agent's observation(s) at any time in the game is the text information perceived by the agent. A probability function takes in the environment state and selects what information to show the agent based on the command entered. The agent receives points based on completion of (sub)quests and reaching new locations (exploring). This score could be used as the reward signal if it is available, otherwise positive reward signals can be assigned when the agent finishes the game. The agent's policy maps the state of the environment at any time and words generated in the command so far to the next word, which needs to be added to the command to maximise the reward received.

\subsubsection{Benchmark tasks}

TextWorld was introduced with two different sets of benchmark tasks \cite{TextWorld} and a third task was added in the form of a competition that was available until 31 May 2019.

\textbf{Task 1:} A preliminary set of 50 hand-authored benchmark games are described in the original TextWorld paper \cite{TextWorld}. These games were manually analysed to ensure validity.

\textbf{Task 2:} This benchmark task is inspired by a treasure hunter task which takes place in a 3D environment  \cite{parisotto2017neural} and was adapted for TextWorld. The agent is randomly placed in a randomly generated map of rooms with two objects on the map. The goal object (the object which the agent should locate) is randomly selected and is mentioned in the welcome message. In order to navigate the map and locate the goal object, the agent may need to complete other tasks, for example finding a key to unlock a cabinet.

This task assesses the agent's skills of affordance extraction, efficient navigation and memory. There are different levels for the benchmark, ranging from level 1 to 30, with different difficulty modes, number of rooms and quest length. 

\textbf{Task 3:} The TextWorld environment is still very new: TextWorld was only released to the public in July 2018. A competition -- First TextWorld Problems: A Reinforcement and Language Learning Challenge\footref{Textworld_url}, which ran until 31 May 2019, was launched by Microsoft Research Montreal to challenge researchers to develop agents that can solve these text-based games. The challenge is gathering ingredients to cook a recipe. 

Agents must determine the necessary ingredients from a recipe book, explore the house to gather ingredients, and return to the kitchen to cook up a delicious meal.

\subsubsection{Published benchmark results}

C{\^{o}}t{\'{e}} \textit{et al.} \cite{TextWorld} evaluate three baseline agents on the benchmark set in Task 1: BYU, Golovin and Simple. The BYU\footnote{\url{https://github.com/danielricks/BYU-Agent-2016}} agent \cite{fulda2017can} utilises a variant of Q-learning \cite{watkins1992q} where word embeddings are trained to be aware of verb-noun affordances. The agent won the IEEE CIG Text-based adventure AI Competition in 2016. The Golovin\footnote{\url{https://github.com/Kostero/text_rpg_ai}} agent \cite{kostka2017text} was developed specifically for classic text-based games and uses a language model pre-trained on fantasy books to extract important keywords from scene descriptions. The Simple\footnote{\url{https://github.com/Microsoft/TextWorld/tree/master/notebooks}} agent uniformly samples a command from a predefined set at every step. Results indicated that all three baseline agents achieved low scores in the games. This indicates that there is significant scope for algorithms to improve on these results.

C{\^{o}}t{\'{e}} \textit{et al.} \cite{TextWorld} also provide average performance results of three agents (BYU, Golovin and a random agent) on 100 treasure hunter games (task 2) at different levels of difficulty. On difficulty level 1 the Golovin agents had the best average score, but the Random agent completed the game in the least number of steps. As the level of difficulty increase, the Random agent achieved the best score and also completed the game in the least number of steps. These results can be used as a baseline for evaluating improved algorithms.

It is evident that there is still enormous scope for research in the environment of text-based games, and that the generative functionality of the TextWorld sandbox environment is a significant contribution in the endeavour of researchers trying to solve these problems.

\subsection{Summary}

For the reader's convenience a summary of the discussed frameworks and algorithms that were shown to be effective are presented in Table~\ref{Table1:Summary}. It should be noted that since the field moves at a rapid pace, the current state of the art will change (it may also be problem instance dependent within the benchmark class), however the listed algorithms can serve as a reasonable baseline for future research. 

\begin{table}[h!]
\begin{center}
\begin{tabular}{ |m{4cm}|m{5cm}|m{7cm}| } 
 \hline
 \textbf{Framework} & \textbf{Benchmark class} & \textbf{Recent effective RL algorithm(s)}  \\ \hline \hline
  OpenAI Gym & Algorithmic & UREX \cite{UREX} \\ \hline

  ~ & Box2D & REINFORCE \cite{Ha_paper} \\ \hline
 ~ & Classic control & TNPG and TRPO \cite{duan2016benchmarking} \\ \hline
~ & MuJoCo & PPO \cite{PPO_17} \\ \hline
~ & Roboschool & PPO \cite{PPO_17} \\ \hline
~ & Robotics & HER \cite{andrychowicz2017hindsight}\footref{Request_Robotics} \\ \hline
 ~ & Toy text & BIRL \cite{cundy2018exploring} \\ \hline \hline
The ALE & Atari 2600 & A2C, ACER and PPO \cite{PPO_17}; A3C \cite{mnih2016asynchronous}; Distribution DQN, Dueling DDQN, Prioritized DDQN and Rainbow \cite{hessel2018rainbow} \tablefootnote{A table summarising the best performance per game can be found at \url{https://github.com/cshenton/atari-leaderboard}.} \\ \hline
 \hline
Garage & Basic tasks & TNPG and TRPO \cite{duan2016benchmarking}  \\ \hline
~ & Locomotion tasks & PPO \cite{PPO_17}  \\ \hline
~ & Partially observable tasks & TNPG and TRPO \cite{duan2016benchmarking}  \\ \hline
~ & Hierarchical tasks & HIRO \cite{NIPS2018_7591}  \\ \hline
\hline
 Keepaway soccer & Keepaway & Episodic SMDP SARSA($\lambda$) \cite{stone2005reinforcement, SuttonBarto_RLI_98} \\ \hline
  ~ & Half-Field Offence & SARSA \cite{hausknecht2016half} \\ \hline \hline
  TextWorld & Original tasks 1, 2 and 3 & BYU and Golovin \cite{TextWorld}  \\ \hline
 ~ & Generalisation tasks & GATA \cite{adhikari2020learning}  \\ \hline
\end{tabular}
\end{center}
\caption{A summary of recent algorithms that performed well in different benchmark sets.}
\label{Table1:Summary}
\end{table}


\section{Discussion}\label{Sect:Summary}

This section focuses on the ways that the different RL benchmarks discussed in Section~\ref{Sect:Contributions} deal with or facilitate research in addressing the challenges for RL discussed in Section~\ref{Subsect:Challenges}.

\subsection{Partially observable environment}

In many of the benchmark tasks, such as the classic control tasks in OpenAI Gym, the agent is provided with full information of the environment. The environment in TextWorld games, however, is partially observable since only local information and the player's inventory are available. The agent might also not be able to distinguish between some states based on observations if only the latest observation is taken into account, i.e. knowledge of past observations are important. In TextWorld games the environment might provide the same feedback for different commands and some important information about certain aspects of the environment might not be available by a single observation. Additionally, the agent might encounter observations that are time-sensitive, such as only being rewarded when it first examines a clue but not any other time. Controlling the partial observability of the state is also part of TextWorld's generative functionality. This is done by augmenting the agent’s observations, where the agent can be provided with a list of present objects or even all game state information can be provided.

The partially observable tasks introduced in rllab (see Section~\ref{rllab:Tasks}), provide environments to investigate agents developed for dealing with environments where not all the information is known.

In RoboCup, a player can by default only observe objects in a 90-degree cone in front of them. In works from Kuhlmann and Stone \cite{kuhlmann2003progress} and Stone \textit{et al.} \cite{stone2005reinforcement} it was shown that it is possible for learning to occur in this limited vision scenario, however players do not perform at an adequate level. For this reason, players in the standardised Keepaway task \cite{stone2005keepaway} operate with 360-vision. 

\subsection{Delayed or sparse rewards}

The tasks in the ALE and TextWorld are interesting when considering reward structure. In the ALE, reward or feedback may only be seen after thousands of actions. In TextWorld, the agent has to generate a sequence of actions before any change in the environment might occur or a reward is received. This results in sparse and delayed rewards in the games, in cases where an agent could receive a positive reward only after many steps when following an optimal strategy. In Keepaway, there is immediate reward, since the learners receive a positive reward after each action they execute.

\subsection{Unspecified or multi-objective reward functions}

In HFO (Section~\ref{HFO}) success not only includes maintaining possession of the ball (the main objective in Keepaway), but the offense players also need to learn to pass or dribble to move towards the goal and shoot when an angle is open. Moreover, success is only evaluated based on a scored goal at the end of an episode, which is rare initially. This aspect of HFO could serve as an ideal environment for investigation into the challenge of problems with multi-objectives.  

Due the definition of a quest in TextWorld, i.e. a sequence of actions where each action depends on the outcomes of the previous action, quests in TextWorld are limited to simple quests. However, in text adventure games, quests are often more complicated, involving multiple sub-quests. C{\^{o}}t{\'{e}} \textit{et al.} \cite{TextWorld} remark that this limitation could be overcome by treating a quest as a directed graph of dependent actions rather than a linear chain. If this can be incorporated in TextWorld in the future, the platform can also be used to study problems with multi-objectives and rewards of varying difficulty.

\subsection{Size of the state and action spaces}

The benchmark tasks that are considered in this paper are ideal to investigate how the size of the state and/or action space challenge can be addressed. The tasks considered all have continuous or large discrete state spaces.

In the ALE the number of states in the games are very large and in TextWorld the state space is combinatorially enormous; since the number of possible states increases exponentially with the number of rooms and objects \cite{TextWorld}. In most of the tasks in OpenAI Gym, rllab, and in Keepaway, the state space is continuous. In Keepaway, the size of the Keepaway region can be varied along with the number of keepers and takers. This allows for investigation into a problem with various difficulties due to the size of the state space.

In TextWorld, the action space is large and sparse because the set of all possible word strings is much larger than the subset of valid commands.  TextWorld's generative functionality also allows control over the size of the state space, i.e. the number of rooms, objects and commands. Different problem difficulties can therefore arise in terms of the size of the state space and this can aid in the investigation of algorithm behaviour with increasing state and action spaces.

\subsection{The trade-off between exploration and exploitation}

In the ALE the challenge of exploration vs. exploitation is difficult due to the large state spaces of games and delayed reward. Simple agents sometimes even learn that staying put is the best policy, since exploration can in some cases lead to negative rewards. Recently there has been some effort to address the exploration problem in the ALE, but these efforts are mostly successful only in individual games. 

Exploration is fundamental to TextWorld games as solving them can not be done by learning a purely exploitative or reactive agent. The agent must use directed exploration as its strategy, where it collects information about objects it encounters along the way. This information will provide knowledge about the goal of the game and provide insight into the environment and what might be useful later in the game. Due to this, exploration by curiosity driven agents might fair well in these types of problems.

Overall, there is still much work to be done to try and overcome this difficult challenge. Machado \textit{et al.} \cite{machado2018revisiting} suggest a few approaches for the ALE, such as agents capable of exploring in a more abstract manner (akin to humans) and agents not exploring joystick movements, but rather exploring object configurations and game levels. Agents with some form of intrinsic motivation might also be needed in order to continue playing even though achieving any reward might seem impossible.

\subsection{Representation learning}

The original goal of the ALE was to develop agents capable of generalising over many games making it desirable to automatically learn representations instead of hand crafting features. Deep Q-Networks (DQN) \cite{mnih2015human} and DQN-like approaches are currently the best overall performing methods, despite high sample complexity. However, additional tuning is often required to obtain better performance \cite{islam2017reproducibility}, which suggest that there is still work to be done to improve performance by learning better representation in the ALE. Other different approaches and directions for representation learning that have been used in the literature are also mentioned in \cite{machado2018revisiting} and should still be explored more in the ALE.

\subsection{Transfer learning}

Regarding the ALE, many of the Atari 2600 games have similar game dynamics and knowledge transfer should reduce the number of samples that are required to learn to play games that are similar. Even more challenging would be determining how to use general video game experience and share that knowledge across games that are not necessarily similar. Current approaches in the literature that apply transfer learning in the ALE are restricted to only a limited subset of games that share similarities and the approaches are based on using neural networks to perform transfer, combining representations and policy transfer. Machado \textit{et al.} \cite{machado2018revisiting} point out that it might be interesting to determine whether transferring each of these entities independently could be helpful. To help with the topic of transfer learning in the ALE, the new version includes different game modes and difficulty settings called flavours (see Section~\ref{Sect:ALE}), which introduces many new environments that are very similar.

Some of the tasks in rllab and environments in OpenAI Gym have been used in studying the transferring of system dynamics from simulation to robots \cite{held2017probabilistically, wulfmeier2017mutual, peng2018sim}. These simulation tasks are an ideal way to safely study the transferring of policies for robotic domains.

Transfer learning has also been studied in the Keepaway soccer domain \cite{taylor2005behavior}, which is a fitting setting since the number of players as well as the size of the action and state spaces can differ.

TextWorld's generative functionality (described in full in \cite{TextWorld}) allows for control of the size and the partial observability of the state space, and therefore a large number of games with shared characteristics can be generated. This could be used for studying transfer learning in text-based games, since agents can be trained on simpler tasks and behaviour transferred to harder problems.

\subsection{Model learning}

Planning and model learning in complex domains are challenging problems and little research has been conducted on this topic compared to traditional RL techniques to learn policies or value functions.

In the ALE, the Stella emulator provides a generative model that can be used in planning and the agent has an exact model of the environment. However, there has not been any success with planning using a learned generative model in the ALE, which is a challenging task since errors start to compound after only a few time steps. A few relatively successful approaches \cite{NIPS2015_5859, chiappa2017recurrent} are available, but the models are slower than the emulator. A challenging open problem is to learn a fast and accurate model for the ALE. On the other hand, related to this, is the problem of planning using an imperfect model. 

On tasks in OpenAI Gym and rllab some research has also been conducted in model learning \cite{nagabandi2018neural, wang2019benchmarking}, but the main focus in the literature is on model-free learning techniques. Therefore there is still scope for substantial research to address this problem.

Wang \textit{et al.} \cite{wang2019benchmarking} attempted to address the lack of a standardised benchmarking framework for model-based RL. They benchmarked 11 model-based RL algorithms and four model-free RL algorithms across 18 environments from OpenAI Gym and have shared the code in an online repository\footnote{\url{http://www.cs.toronto.edu/~tingwuwang/mbrl.html}}. They evaluated the efficiency, performance and robustness of three different categories of model-based RL algorithms (Dyna style algorithms, policy search with backpropagation through time and shooting algorithms) and four model-free algorithms (TRPO, PPO, TD3, and SAC -- refer to Section~\ref{Subsect:Model_free_vs_based} for these algorithms).  They also propose three key research challenges for model-based methods, namely the dynamics bottleneck, the planning horizon dilemma, and the early termination dilemma and show that even with substantial benchmarking, there is no clear consistent best model-based RL algorithm. This again suggests that there is substantial scope and many opportunities for further research in model-based RL methods.

\subsection{Off-policy learning}

Deep neural networks have become extremely popular in modern RL literature, and the breakthrough work of Mnih \textit{et al.} \cite{Mnih_et_al_Atari_13, mnih2015human} demonstrates DQN having human-level performance on Atari 2600 games. However, when using deep neural networks for function approximation for off-policy algorithms, new and complex challenges arise, such as instability and slow convergence. While discussing off-policy methods using function approximation, Sutton and Barto \cite{SuttonBarto_RLI_98} conclude the following: ``The potential for off-policy learning remains tantalizing, the best way to achieve it still a mystery." Nevertheless, off-policy learning has become an active research field in RL.

The use of off-policy learning algorithms in the ALE in  current literature varies with most approaches using experience replay and target networks. This is an attempt at reducing divergence in off-policy learning, but these methods are very complex. New proposed algorithms such as GQ($\lambda$) \cite{Maei2010/06} are theoretically sound, but there is still a need for a thorough empirical evaluation or demonstration of these theoretically sound off-policy learning RL algorithms. Other contributions of using off-policy learning in the ALE includes double Q-learning \cite{Hasselt:2016} and Q($\lambda$) with off-policy corrections \cite{Harutyunyan_2016}.

Some of the tasks in rllab and OpenAI Gym have also been used in studying off-policy algorithms, for example introducing the soft actor-critic (SAC) algorithm \cite{haarnoja2018soft} and using the robotics environments from OpenAI Gym to learn grasping \cite{quillen2018deep}. This area of research is still new and there is significant scope for further research in this domain.

\subsection{Reinforcement learning in real-world settings}

The robotics environments in the OpenAI Gym toolkit can be used to train models which work on physical robots. This can be used to develop agents to safely execute realistic tasks. A request for research from OpenAI\footref{Request_Robotics} indicates that work in this area is an active research field with promising results.

The Keepaway and HFO soccer tasks are ideal settings to study multi-agent RL \cite{4445757}, an important research area for real-world problems since humans act in an environment where objectives are shared with others. 

Challenges for RL that are unique to TextWorld games are related to natural language understanding: observation modality, understanding the parser feedback, common-sense reasoning and affordance extraction, and language acquisition. These challenges are explained in more detail in C{\^{o}}t{\'{e}} \textit{et al.} \cite{TextWorld}. Natural language understanding is an important aspect of artificial intelligence, in order for communication to take place between humans and AI. TextWorld can be used to address many of the challenges described in Section~\ref{Subsect:Challenges} in simpler settings and to focus on testing and debugging agents on subsets of these challenges.

In addition to the frameworks covered in this survey, there are two further contributions that are focused on multi-agent and distributed RL. The MAgent research platform \cite{zheng2018magent} facilitates research in many-agent RL, specifically in artificial collective intelligence. The platform aims at supporting RL research that scales up from hundreds to millions of agents and is maintained in an online repository\footnote{\url{https://github.com/geek-ai/MAgent}}. MAgent 
also provides a visual interface presenting the state of the environment and agents.

A research team from Stanford has introduced the open-source framework SURREAL (\textbf{S}calable \textbf{R}obotic \textbf{RE}inforcementlearning
\textbf{AL}gorithms) and the SURREAL Robotics Suite \cite{corl2018surreal}, to facilitae research in RL in robotics and distributed RL. SURREAL eliminates the need for global synchronization and improves scalability by decoupling a distributed RL algorithm into four
components. The four-layer computing infrastructure can easily be deployed on commercial cloud providers or personal computers, and is also fully replicable from scratch, contributing to the reproducibility of results. The Robotics Suite is developed in the MuJoCo physics engine and provides OpenAI gym-style interfaces in Python. Detailed API documentation and tutorials on importing new robots and the creation of new environments and tasks are also provided, furthering the contribution to research in this field. The Robotics Suite is actively maintained in an online repository\footnote{\url{https://github.com/SurrealAI/surreal}}. The different robotics tasks include block lifting and stacking, bimanual peg-in-hole placing and bimanual lifting, bin picking, and nut-and-peg assembly. Variants of PPO and DDPG called SURREAL-PPO and SURREAL-DDPG were developed and examined on the Robotics Suite tasks, and experiments indicate that these SURREAL algorithms can achieve good results.

\subsection{A standard methodology for benchmarking}

The ALE consists of games with similar structure in terms of of inputs, action movements, etc. This makes the ALE an ideal benchmark for comparative studies. A standard methodology is however needed and this is proposed by Machado \textit{et al.} \cite{machado2018revisiting}:
\begin{itemize}
\item Episode termination can be standardised by using the game over signal than lives lost.
\item Hyperparameter tuning needs to be consistently applied on the training set only.
\item Training time should be consistently applied across different problems.
\item There is a need for standard ways of reporting learning performance.
\end{itemize}

These same principles apply to groups of similar tasks in OpenAI Gym and rllab, and to TextWorld and Keepaway soccer.


\subsection{Trends in benchmarking of RL}

It is clear from Section \ref{Sect:Contributions} that the number of well thought-out frameworks designed for RL benchmarks has rapidly expanded in recent years, with a general move to fully open source implementations being evident. A notable example is OpenAI Gym re-implementing, to an extent, open source variants of the benchmarks previously provided in the MuJoCo simulation environment. The move to fully open source implementations has had two primary benefits: reproducibility and accessibility. 

The variety of RL frameworks and benchmark sets may present a challenge to a novice in the field, as there is no clear standard benchmark set or framework to use. This is not a surprising situation as the array of RL application areas has become relatively diverse and so different types of problems and their corresponding challenges will naturally be more interesting to certain sub-communities within the field. 

One aspect of modern RL benchmarks that is relatively striking is the increase in problem complexity. While it is not immediately clear how to precisely define problem difficulty, it is clear that more and more problem features that are challenging for RL algorithms are being included in proposed benchmarks. Many established benchmark sets have been explicitly expanded to increase the challenge of a given problem instance. Some notable examples include the addition of sticky actions in the ALE and the addition of the partially observable variants of rllab's continuous control tasks.

It is also clear that the advancements made in the field of deep learning has allowed for certain types of RL tasks to be more readily solvable. Two notable examples are the use of convolution neural networks \cite{CNNLecun} to assist in the vision problem present in Atari 2600 games of the ALE, and the use of modern neutral network based approaches to natural language processing in Microsoft's TextWorld. 


\section{Conclusion}\label{Sect:Conclusion}

This paper provides a survey of some of the most used and recent contributions to RL benchmarking. A number of benchmarking frameworks are described in terms of their characteristics, technical implementation details and the tasks provided. A summary is also provided of published results on the performance of algorithms used to solve these benchmark tasks. Challenges that occur when solving RL problems are also discussed, including the various ways the different benchmarking tasks address or facilitate research in addressing these challenges. 

The survey reveals that there has been substantial progress in the endeavour of standardising benchmarking tasks for RL. The research community has started to acknowledge the importance of reproducible results and research has been published to encourage the community to address this problem. However, there is still a lot to be done in ensuring the reproducibility of results for fair comparison.

There are many approaches when solving RL problems and proper benchmarks are important when comparing old and new approaches. This survey indicates that the tasks currently used for benchmarking RL encompass a wide range of problems and can even be used to develop algorithms for training agents in real-world systems such as robots.

\bibliographystyle{unsrtnat}
\bibliography{bibliography_new}

\begin{thebibliography}{90}
\providecommand{\natexlab}[1]{#1}
\providecommand{\url}[1]{\texttt{#1}}
\expandafter\ifx\csname urlstyle\endcsname\relax
  \providecommand{\doi}[1]{doi: #1}\else
  \providecommand{\doi}{doi: \begingroup \urlstyle{rm}\Url}\fi

\bibitem[Sutton and Barto(2018)]{SuttonBarto_RLI_98}
R.~S. Sutton and A.~G. Barto.
\newblock \emph{Reinforcement Learning: An Introduction}.
\newblock MIT Press, 2018.
\newblock ISBN 978-0262039246.

\bibitem[Lillicrap et~al.(2015)Lillicrap, Hunt, Pritzel, Heess, Erez, Tassa,
  Silver, and Wierstra]{Lillicrap_et_al_15}
T.~P. Lillicrap, J.~J. Hunt, A.~Pritzel, N.~Heess, T.~Erez, Y.~Tassa,
  D.~Silver, and D.~Wierstra.
\newblock Continuous control with deep reinforcement learning.
\newblock arXiv:1509.02971, 2015.

\bibitem[Mnih et~al.(2013)Mnih, Kavukcuoglu, Silver, Graves, Antonoglou,
  Wierstra, and Riedmiller]{Mnih_et_al_Atari_13}
V.~Mnih, K.~Kavukcuoglu, D.~Silver, A.~Graves, A.~Antonoglou, A.~Wierstra, and
  M.~Riedmiller.
\newblock Playing {A}tari with deep reinforcement learning.
\newblock arXiv:1312.5602, 2013.

\bibitem[Mnih et~al.(2015)Mnih, Kavukcuoglu, Silver, Rusu, Veness, Bellemare,
  Graves, Riedmiller, Fidjeland, Ostrovski, Petersen, Beattie, Sadik,
  Antonoglou, King, Kumaran, Wierstra, Legg, and Hassabis]{mnih2015human}
V.~Mnih, K.~Kavukcuoglu, D.~Silver, A.~A. Rusu, J.~Veness, M.~G. Bellemare,
  A.~Graves, M.~Riedmiller, A.~K. Fidjeland, G.~Ostrovski, S.~Petersen,
  C.~Beattie, A.~Sadik, I.~Antonoglou, H.~King, D.~Kumaran, D.~Wierstra,
  S.~Legg, and D.~Hassabis.
\newblock Human-level control through deep reinforcement learning.
\newblock \emph{Nature}, 518:\penalty0 529, 2015.
\newblock \doi{https://doi.org/10.1038/nature14236}.

\bibitem[Vinyals et~al.(2017)Vinyals, Ewalds, Bartunov, Georgiev, Vezhnevets,
  Yeo, Makhzani, Küttler, Agapiou, Schrittwieser, Quan, Gaffney, Petersen,
  Simonyan, Schaul, {van Hasselt}, Silver, Timothy~Lillicrap, Kevin~Calderone,
  Paul~Keet, Brunasso, Lawrence, Ekermo, Repp, and
  Tsing]{Vinyals_et_al_Starcraft_17}
O.~Vinyals, T.~Ewalds, S.~Bartunov, P.~Georgiev, A.~S. Vezhnevets, M.~Yeo,
  A.~Makhzani, H.~Küttler, J.~Agapiou, J.~Schrittwieser, J.~Quan, S.~Gaffney,
  S.~Petersen, K.~Simonyan, T.~Schaul, H.~{van Hasselt}, D.~Silver,
  T.~Timothy~Lillicrap, K.~Kevin~Calderone, P.~Paul~Keet, A.~Brunasso,
  D.~Lawrence, A.~Ekermo, J.~Repp, and R.~Tsing.
\newblock Star{C}raft {II}: {A} new challenge for reinforcement learning.
\newblock arXiv:1708.04782, 2017.

\bibitem[Silva and Chaimowicz(2017)]{Silva_Chaimowicz_Moba_17}
V.~d.~N. Silva and L.~Chaimowicz.
\newblock {MOBA}: {A} new arena for game {AI}.
\newblock arXiv:1705.10443, 2017.

\bibitem[Arel et~al.(2010)Arel, Liu, Urbanik, and Kohls]{Traffic_control}
I.~Arel, C.~Liu, T.~Urbanik, and A.~Kohls.
\newblock Reinforcement learning-based multi-agent system for network traffic
  signal control.
\newblock \emph{IET Intelligent Transport Systems}, 4\penalty0 (2):\penalty0
  128--135, 2010.
\newblock \doi{https://doi.org/10.1049/iet-its.2009.0070}.

\bibitem[Silver et~al.(2016)Silver, Huang, Maddison, Guez, Sifre, {van den
  Driessche}, Schrittwieser, Antonoglou, Panneershelvam, Lanctot, Dieleman,
  Grewe, Nham, Kalchbrenner, Sutskever, Lillicrap, Leach, Kavukcuoglu, Graepel,
  and Hassabis]{Silver_et_al_GO_16}
D.~Silver, A.~Huang, C.~J. Maddison, A.~Guez, L.~Sifre, G.~{van den Driessche},
  J.~Schrittwieser, I.~Antonoglou, V.~Panneershelvam, M.~Lanctot, S.~Dieleman,
  D.~Grewe, J.~Nham, N.~Kalchbrenner, I.~Sutskever, T.~Lillicrap, M.~Leach,
  K.~Kavukcuoglu, T.~Graepel, and D.~Hassabis.
\newblock Mastering the game of {G}o with deep neural networks and tree search.
\newblock \emph{Nature}, 529:\penalty0 484--489, 2016.
\newblock \doi{https://doi.org/10.1038/nature16961}.

\bibitem[Marcus et~al.(1993)Marcus, Santorini, and
  Marcinkiewicz]{marcus1993building}
M.~Marcus, B.~Santorini, and M.~A. Marcinkiewicz.
\newblock Building a large annotated corpus of {English}: {The Penn Treebank}.
\newblock \emph{Computational Linguistics}, 19\penalty0 (2):\penalty0 313--330,
  jun 1993.
\newblock ISSN 0891-2017.

\bibitem[Russakovsky et~al.(2015)Russakovsky, Deng, Su, Krause, Satheesh, Ma,
  Huang, Karpathy, Khosla, Bernstein, Berg, and Fei-Fei]{Russakovsky2015}
O.~Russakovsky, J.~Deng, H.~Su, J.~Krause, S.~Satheesh, S.~Ma, Z.~Huang,
  A.~Karpathy, A.~Khosla, M.~Bernstein, A.~C. Berg, and L.~Fei-Fei.
\newblock Imagenet large scale visual recognition challenge.
\newblock \emph{International Journal of Computer Vision}, 115\penalty0
  (3):\penalty0 211--252, Dec 2015.
\newblock ISSN 1573-1405.
\newblock \doi{https://doi.org/10.1007/s11263-015-0816-y}.

\bibitem[Everingham et~al.(2010)Everingham, Van~Gool, Williams, Winn, and
  Zisserman]{Everingham2010}
M.~Everingham, L.~Van~Gool, C.~K.~I. Williams, J.~Winn, and A.~Zisserman.
\newblock {The Pascal Visual Object Classes (VOC) Challenge}.
\newblock \emph{International Journal of Computer Vision}, 88\penalty0
  (2):\penalty0 303--338, Jun 2010.
\newblock ISSN 1573-1405.
\newblock \doi{https://doi.org/10.1007/s11263-009-0275-4}.

\bibitem[Bellemare et~al.(2013)Bellemare, Naddaf, Veness, and Bowling]{ALE}
M.~G. Bellemare, Y.~Naddaf, J.~Veness, and M.~Bowling.
\newblock The {A}rcade {L}earning {E}nvironment: {A}n evaluation platform for
  general agents.
\newblock \emph{Journal of Artificial Intelligence Research}, 47:\penalty0
  253--279, 2013.
\newblock ISSN 1076-9757.
\newblock \doi{https://doi.org/10.1613/jair.3912}.

\bibitem[Duan et~al.(2016)Duan, Chen, Houthooft, Schulman, and
  Abbeel]{duan2016benchmarking}
Y.~Duan, X.~Chen, R.~Houthooft, J.~Schulman, and P.~Abbeel.
\newblock Benchmarking deep reinforcement learning for continuous control.
\newblock In \emph{Proceedings of the 33rd International Conference on
  International Conference on Machine Learning}, ICML'16, pages 1329--1338,
  2016.

\bibitem[Henderson et~al.(2018)Henderson, Islam, Bachman, Pineau, Precup, and
  Meger]{Henderson_et_al_18}
P.~Henderson, R.~Islam, P.~Bachman, J.~Pineau, D.~Precup, and D.~Meger.
\newblock Deep reinforcement learning that matters.
\newblock In \emph{{P}roceedings of the {T}hirty-{S}econd {AAAI} Conference on
  Artificial Intelligence}, pages 3207--3214, 2018.

\bibitem[Machado et~al.(2018)Machado, Bellemare, Talvitie, Veness, Hausknecht,
  and Bowling]{machado2018revisiting}
M.~C. Machado, M.~G. Bellemare, E.~Talvitie, J.~Veness, M.~Hausknecht, and
  M.~Bowling.
\newblock Revisiting the {A}rcade {L}earning {E}nvironment: Evaluation
  protocols and open problems for general agents.
\newblock \emph{Journal of Artificial Intelligence Research}, 61:\penalty0
  523--562, 2018.
\newblock ISSN 1076-9757.

\bibitem[Brockman et~al.(2016)Brockman, Cheung, Pettersson, Schneider,
  Schulman, Tang, and Zaremba]{OpenAI_Gym}
G.~Brockman, V.~Cheung, L.~Pettersson, J.~Schneider, J.~Schulman, J.~Tang, and
  W.~Zaremba.
\newblock {OpenAI} {G}ym.
\newblock arXiv:1606.01540, 2016.

\bibitem[Stone and Sutton(2001)]{stone2001keepaway}
P.~Stone and R.~S. Sutton.
\newblock Keepaway soccer: A machine learning test bed.
\newblock In \emph{Robot Soccer World Cup}, pages 214--223. Springer, 2001.
\newblock \doi{https://doi.org/10.1007/11780519_9}.

\bibitem[C{\^{o}}t{\'{e}} et~al.(2018)C{\^{o}}t{\'{e}}, K{\'{a}}d{\'{a}}r,
  Yuan, Kybartas, Barnes, Fine, Moore, Hausknecht, Asri, Adada, Tay, and
  Trischler]{TextWorld}
M.~A. C{\^{o}}t{\'{e}}, {\'{A}}~K{\'{a}}d{\'{a}}r, X.~Yuan, B.~Kybartas,
  T.~Barnes, E.~Fine, J.~Moore, M.~J. Hausknecht, L.~E. Asri, M.~Adada, W.~Tay,
  and A.~Trischler.
\newblock {TextWorld}: {A} learning environment for text-based games.
\newblock arXiv:1806.11532, 2018.

\bibitem[Mnih et~al.(2016)Mnih, Badia, Mirza, Graves, Lillicrap, Harley,
  Silver, and Kavukcuoglu]{mnih2016asynchronous}
V.~Mnih, A.~P. Badia, M.~Mirza, A.~Graves, T.~Lillicrap, T.~Harley, D.~Silver,
  and K.~Kavukcuoglu.
\newblock Asynchronous methods for deep reinforcement learning.
\newblock In \emph{Proceedings of the 33rd International Conference on
  International Conference on Machine Learning}, ICML'16, pages 1928--1937,
  2016.

\bibitem[Schulman et~al.(2015)Schulman, Levine, Moritz, Jordan, and
  Abbeel]{Schulman_etal_2015}
J.~Schulman, S.~Levine, P.~Moritz, M.~Jordan, and P.~Abbeel.
\newblock Trust region policy optimization.
\newblock In \emph{Proceedings of the 32nd International Conference on Machine
  Learning}, ICML'15, pages 1889--1897, 2015.

\bibitem[Schulman et~al.(2017)Schulman, Wolski, Dhariwal, Radford, and
  Klimov]{PPO_17}
J.~Schulman, F.~Wolski, P.~Dhariwal, A.~Radford, and O.~Klimov.
\newblock Proximal policy optimization algorithms.
\newblock arXiv:1707.06347, 2017.

\bibitem[Bellemare et~al.(2017)Bellemare, Dabney, and
  Munos]{bellemare2017distributional}
M.~G. Bellemare, W.~Dabney, and R.~Munos.
\newblock A distributional perspective on reinforcement learning.
\newblock In \emph{Proceedings of the 34th International Conference on Machine
  Learning}, ICML'17, pages 449--458, 2017.

\bibitem[Andrychowicz et~al.(2017)Andrychowicz, Wolski, Ray, Schneider, Fong,
  Welinder, McGrew, Tobin, Abbeel, and Zaremba]{andrychowicz2017hindsight}
M.~Andrychowicz, F.~Wolski, A.~Ray, J.~Schneider, R.~Fong, P.~Welinder,
  B.~McGrew, J.~Tobin, P.~Abbeel, and W.~Zaremba.
\newblock Hindsight experience replay.
\newblock In \emph{Advances in Neural Information Processing Systems 30}, pages
  5048--5058, 2017.

\bibitem[Haarnoja et~al.(2018)Haarnoja, Zhou, Abbeel, and
  Levine]{haarnoja2018soft}
T.~Haarnoja, A.~Zhou, P.~Abbeel, and S.~Levine.
\newblock Soft actor-critic: {O}ff-policy maximum entropy deep reinforcement
  learning with a stochastic actor.
\newblock In \emph{Proceedings of the 35th International Conference on Machine
  Learning}, ICML'18, pages 1861--1870, 2018.

\bibitem[Fujimoto et~al.(2018)Fujimoto, van Hoof, and
  Meger]{fujimoto2018addressing}
S.~Fujimoto, H.~van Hoof, and D.~Meger.
\newblock Addressing function approximation error in actor-critic methods.
\newblock In \emph{Proceedings of the 35th International Conference on Machine
  Learning}, ICML'18, pages 1587--1596, 2018.

\bibitem[Silver et~al.(2012)Silver, Sutton, and M{\"u}ller]{Silver2012}
D.~Silver, R.~S. Sutton, and M.~M{\"u}ller.
\newblock Temporal-difference search in computer {Go}.
\newblock \emph{Machine Learning}, 87\penalty0 (2):\penalty0 183--219, 2012.
\newblock ISSN 1573-0565.
\newblock \doi{https://doi.org/10.1007/s10994-012-5280-0}.

\bibitem[Minsky(1961)]{minsky1961steps}
M.~Minsky.
\newblock Steps toward artificial intelligence.
\newblock \emph{Proceedings of the IRE}, 49\penalty0 (1):\penalty0 8--30, 1961.
\newblock \doi{https://doi.org/10.1109/JRPROC.1961.287775}.

\bibitem[Bellman(1957)]{Bellman_1957}
R.~E. Bellman.
\newblock \emph{Dynamic Programming}.
\newblock Princeton University Press, Princeton, 1957.
\newblock ISBN 978-0486428093.

\bibitem[{Pan} and {Yang}(2010)]{Pan_Yang_Transfer_2010}
S.~J. {Pan} and Q.~{Yang}.
\newblock A survey on transfer learning.
\newblock \emph{IEEE Transactions on Knowledge and Data Engineering},
  22\penalty0 (10):\penalty0 1345--1359, 2010.
\newblock ISSN 1041-4347.
\newblock \doi{https://doi.org/10.1109/TKDE.2009.191}.

\bibitem[Weiss et~al.(2016)Weiss, Khoshgoftaar, and Wang]{weiss2016survey}
K.~Weiss, T.~M. Khoshgoftaar, and D.~Wang.
\newblock A survey of transfer learning.
\newblock \emph{Journal of Big data}, 3\penalty0 (9), 2016.
\newblock ISSN 2196-1115.
\newblock \doi{https://doi.org/10.1186/s40537-016-0043-6}.

\bibitem[Taylor and Stone(2009)]{taylor2009transfer}
M.~E. Taylor and P.~Stone.
\newblock Transfer learning for reinforcement learning domains: {A} survey.
\newblock \emph{Journal of Machine Learning Research}, 10\penalty0
  (Jul):\penalty0 1633--1685, 2009.
\newblock ISSN 1532-4435.

\bibitem[Dulac-Arnold et~al.(2019)Dulac-Arnold, Mankowitz, and
  Hester]{Real_world_RL_challenges}
G.~Dulac-Arnold, D.~Mankowitz, and T.~Hester.
\newblock Challenges of real-world reinforcement learning.
\newblock In \emph{Reinforcement Learning for Real Life (RL4RealLife) Workshop
  in the 36th International Conference on Machine Learning}, 2019.
\newblock arXiv:1904.12901.

\bibitem[{Todorov} et~al.(2012){Todorov}, {Erez}, and {Tassa}]{MuJoCo_citation}
E.~{Todorov}, T.~{Erez}, and Y.~{Tassa}.
\newblock {MuJoCo}: {A} physics engine for model-based control.
\newblock In \emph{2012 IEEE/RSJ International Conference on Intelligent Robots
  and Systems}, pages 5026--5033, 2012.
\newblock \doi{https://doi.org/10.1109/IROS.2012.6386109}.

\bibitem[Liang et~al.(2016)Liang, Machado, Talvitie, and
  Bowling]{liang2016state}
Y.~Liang, M.~C. Machado, E.~Talvitie, and M.~Bowling.
\newblock State of the art control of {A}tari games using shallow reinforcement
  learning.
\newblock In \emph{Proceedings of the 2016 International Conference on
  Autonomous Agents and Multiagent Systems}, pages 485--493. International
  Foundation for Autonomous Agents and Multiagent Systems, 2016.

\bibitem[Dann et~al.(2014)Dann, Neumann, and Peters]{JMLR:v15:dann14a}
C.~Dann, G.~Neumann, and J.~Peters.
\newblock Policy evaluation with temporal differences: A survey and comparison.
\newblock \emph{Journal of Machine Learning Research}, 15:\penalty0 809--883,
  2014.
\newblock ISSN 1532-4435.

\bibitem[Papis and Wawrzy{\'n}ski(2013)]{papis2013dotrl}
B.~Papis and P.~Wawrzy{\'n}ski.
\newblock dot{RL}: {A} platform for rapid reinforcement learning methods
  development and validation.
\newblock In \emph{2013 Federated Conference on Computer Science and
  Information Systems}, pages 129--136. IEEE, 2013.

\bibitem[Woolley and Stanley(2010)]{woolley_stanley_octupus_arm_2010}
B.~G. Woolley and K.~O. Stanley.
\newblock Evolving a single scalable controller for an octopus arm with a
  variable number of segments.
\newblock In \emph{Parallel Problem Solving from Nature, PPSN XI}, pages
  270--279, Berlin, Heidelberg, 2010. Springer Berlin Heidelberg.
\newblock \doi{https://doi.org/10.1007/978-3-642-15871-1_28}.

\bibitem[Schaul et~al.(2010)Schaul, Bayer, Wierstra, Sun, Felder, Sehnke,
  R\"{u}ckstie\ss, and Schmidhuber]{schaul2010pybrain}
T.~Schaul, J.~Bayer, D.~Wierstra, Y.~Sun, M.~Felder, F.~Sehnke,
  T.~R\"{u}ckstie\ss, and J.~Schmidhuber.
\newblock Py{B}rain.
\newblock \emph{Journal of Machine Learning Research}, 11\penalty0
  (Feb):\penalty0 743--746, 2010.
\newblock ISSN 1532-4435.

\bibitem[Yamaguchi and Ogasawara(2010)]{yamaguchi2010skyai}
A.~Yamaguchi and T.~Ogasawara.
\newblock {SkyAI}: {H}ighly modularized reinforcement learning library.
\newblock In \emph{2010 10th IEEE-RAS International Conference on Humanoid
  Robots}, pages 118--123. IEEE, 2010.
\newblock \doi{https://doi.org/10.1109/ICHR.2010.5686285}.

\bibitem[Williams(1992)]{Williams92_REINFORCE}
R.J. Williams.
\newblock Simple statistical gradient-following algorithms for connectionist
  reinforcement learning.
\newblock \emph{Machine Learning}, 8:\penalty0 229, 1992.
\newblock ISSN 0885-6125.
\newblock \doi{https://doi.org/10.1007/BF00992696}.

\bibitem[Peters and Schaal(2007)]{Peters_Schaal_07}
J.~Peters and S.~Schaal.
\newblock Reinforcement learning by reward-weighted regression for operational
  space control.
\newblock In \emph{Proceedings of the 24th International Conference on Machine
  Learning}, ICML '07, pages 745--750, 2007.
\newblock \doi{https://doi.org/10.1145/1273496.1273590}.

\bibitem[Peters et~al.(2010)Peters, Mulling, and Altun]{AAAI101851}
J.~Peters, K.~Mulling, and Y.~Altun.
\newblock Relative entropy policy search.
\newblock In \emph{Proceedings of the Twenty-Fourth AAAI Conference on
  Artificial Intelligence}, AAAI'10, pages 1607--1612, 2010.

\bibitem[Rubinstein(1999)]{Rubinstein1999}
R.~Rubinstein.
\newblock The cross-entropy method for combinatorial and continuous
  optimization.
\newblock \emph{Methodology And Computing In Applied Probability}, 1\penalty0
  (2):\penalty0 127--190, 1999.
\newblock ISSN 1573-7713.
\newblock \doi{https://doi.org/10.1023/A:1010091220143}.

\bibitem[Hansen and Ostermeier(2001)]{Hansen_Ostermeier_2001}
N.~Hansen and A.~Ostermeier.
\newblock Completely derandomized self-adaptation in evolution strategies.
\newblock \emph{Evolutionary Computation}, 9\penalty0 (2):\penalty0 159--195,
  2001.
\newblock ISSN 1063-6560.
\newblock \doi{https://doi.org/10.1162/106365601750190398}.

\bibitem[Kitano et~al.(1997)Kitano, Asada, Kuniyoshi, Noda, and
  Osawa]{RoboCup_97}
H.~Kitano, M.~Asada, Y.~Kuniyoshi, I.~Noda, and E.~Osawa.
\newblock {RoboCup}: {The} {Robot} {World} {Cup} {Initiative}.
\newblock In \emph{Proceedings of the First International Conference on
  Autonomous Agents}, AGENTS '97, pages 340--347, 1997.
\newblock \doi{https://doi.org/10.1145/267658.267738}.

\bibitem[Stone et~al.(2005{\natexlab{a}})Stone, Sutton, and
  Kuhlmann]{stone2005reinforcement}
P.~Stone, R.~S. Sutton, and G.~Kuhlmann.
\newblock Reinforcement learning for {R}obo{C}up soccer {K}eepaway.
\newblock \emph{Adaptive Behavior}, 13\penalty0 (3):\penalty0 165--188,
  2005{\natexlab{a}}.
\newblock ISSN 1059-7123.
\newblock \doi{https://doi.org/10.1177/105971230501300301}.

\bibitem[Pietro et~al.(2002)Pietro, While, and Barone]{pietro2002learning}
A.~D. Pietro, L.~While, and L.~Barone.
\newblock Learning in {R}obo{C}up keepaway using evolutionary algorithms.
\newblock In \emph{Proceedings of the 4th Annual Conference on Genetic and
  Evolutionary Computation}, GECCO'02, pages 1065--1072, 2002.

\bibitem[Walker et~al.(2004)Walker, Shavlik, and Maclin]{walker2004relational}
T.~Walker, J.~Shavlik, and R.~Maclin.
\newblock Relational reinforcement learning via sampling the space of
  first-order conjunctive features.
\newblock In \emph{Proceedings of the ICML Workshop on Relational Reinforcement
  Learning}, 2004.

\bibitem[Taylor and Stone(2005)]{taylor2005behavior}
M.~E. Taylor and P.~Stone.
\newblock Behavior transfer for value-function-based reinforcement learning.
\newblock In \emph{Proceedings of the fourth international joint conference on
  Autonomous agents and multiagent systems}, AAMAS '05, pages 53--59, 2005.
\newblock \doi{https://doi.org/10.1145/1082473.1082482}.

\bibitem[Didi and Nitschke(2016{\natexlab{a}})]{didi2016multi}
S.~Didi and G.~Nitschke.
\newblock Multi-agent behavior-based policy transfer.
\newblock In \emph{European Conference on the Applications of Evolutionary
  Computation}, EvoApplications, pages 181--197, 2016{\natexlab{a}}.
\newblock \doi{https://doi.org/10.1007/978-3-319-31153-1_13}.

\bibitem[Didi and Nitschke(2016{\natexlab{b}})]{didi2016hybridizing}
S.~Didi and G.~Nitschke.
\newblock Hybridizing novelty search for transfer learning.
\newblock In \emph{IEEE Symposium Series on Computational Intelligence (SSCI)},
  pages 1--8, 2016{\natexlab{b}}.
\newblock \doi{https://doi.org/10.1109/SSCI.2016.7850180}.

\bibitem[Nitschke and Didi(2017)]{nitschke2017evolutionary}
G.~Nitschke and S.~Didi.
\newblock Evolutionary policy transfer and search methods for boosting behavior
  quality: Robocup keep-away case study.
\newblock \emph{Frontiers in Robotics and AI}, 4:\penalty0 62, 2017.
\newblock ISSN 2296-9144.
\newblock \doi{https://doi.org/10.3389/frobt.2017.00062}.

\bibitem[Didi and Nitschke(2018)]{Didi:2018:robocup}
S.~Didi and G.~Nitschke.
\newblock Policy transfer methods in {R}obo{C}up keep-away.
\newblock In \emph{Proceedings of the Genetic and Evolutionary Computation
  Conference Companion}, GECCO '18, pages 117--118, 2018.
\newblock ISBN 978-1-4503-5764-7.
\newblock \doi{https://doi.org/10.1145/3205651.3205710}.

\bibitem[Schwab et~al.(2018)Schwab, Zhu, and
  Veloso]{Schwab:2018:transfer_robocup}
D.~Schwab, Y.~Zhu, and M.~Veloso.
\newblock Zero shot transfer learning for robot soccer.
\newblock In \emph{Proceedings of the 17th International Conference on
  Autonomous Agents and MultiAgent Systems}, AAMAS'18, pages 2070--2072, 2018.

\bibitem[Cheng et~al.(2018)Cheng, Wang, Niu, and Shen]{sym11010025}
Q.~Cheng, X.~Wang, Y.~Niu, and L.~Shen.
\newblock Reusing source task knowledge via transfer approximator in
  reinforcement transfer learning.
\newblock \emph{Symmetry}, 11\penalty0 (1), 2018.
\newblock ISSN 2073-8994.
\newblock \doi{https://doi.org/10.3390/sym11010025}.

\bibitem[Riedmiller et~al.(2009)Riedmiller, Gabel, Hafner, and
  Lange]{riedmiller2009reinforcement}
M.~Riedmiller, T.~Gabel, R.~Hafner, and S.~Lange.
\newblock Reinforcement learning for robot soccer.
\newblock \emph{Autonomous Robots}, 27\penalty0 (1):\penalty0 55--73, 2009.
\newblock \doi{https://doi.org/10.1007/s10514-009-9120-4}.

\bibitem[Bai and Russell(2017)]{bai2017efficient}
A.~Bai and S.~Russell.
\newblock Efficient reinforcement learning with hierarchies of machines by
  leveraging internal transitions.
\newblock In \emph{Proceedings of the Twenty-Sixth International Joint
  Conference on Artificial Intelligence}, pages 1418--1424, 2017.
\newblock \doi{https://doi.org/10.24963/ijcai.2017/196}.

\bibitem[Stone et~al.(2005{\natexlab{b}})Stone, Kuhlmann, Taylor, and
  Liu]{stone2005keepaway}
P.~Stone, G.~Kuhlmann, M.~E. Taylor, and Y.~Liu.
\newblock Keepaway soccer: From machine learning testbed to benchmark.
\newblock In \emph{Robot Soccer World Cup}, pages 93--105, 2005{\natexlab{b}}.
\newblock \doi{https://doi.org/10.1007/11780519_9}.

\bibitem[Stone(2000)]{stone2000layered}
P.~Stone.
\newblock \emph{Layered learning in multiagent systems: A winning approach to
  robotic soccer}.
\newblock MIT Press, 2000.
\newblock ISBN 978-0819428448.

\bibitem[Albus(1975)]{albus1975new}
J.~S. Albus.
\newblock A new approach to manipulator control: {The} cerebellar model
  articulation controller {(CMAC)}.
\newblock \emph{Journal of Dynamic Systems, Measurement, and Control},
  97\penalty0 (3):\penalty0 220--227, 1975.
\newblock \doi{https://doi.org/10.1115/1.3426922}.

\bibitem[Albus(1981)]{Albus_1981}
J.~S. Albus.
\newblock \emph{Brains, {Behavior} and {Robotics}}.
\newblock McGraw-Hill, Inc., 1981.
\newblock ISBN 0070009759.

\bibitem[Kalyanakrishnan et~al.(2007)Kalyanakrishnan, Liu, and Stone]{HFO_2007}
S.~Kalyanakrishnan, Y.~Liu, and P.~Stone.
\newblock {Half Field Offense in RoboCup Soccer: A Multiagent Reinforcement
  Learning Case Study}.
\newblock In \emph{RoboCup 2006: Robot Soccer World Cup X}, pages 72--85, 2007.
\newblock \doi{https://doi.org/10.1007/978-3-540-74024-7_7}.

\bibitem[Hausknecht et~al.(2016)Hausknecht, Mupparaju, Subramanian,
  Kalyanakrishnan, and Stone]{hausknecht2016half}
M.~Hausknecht, P.~Mupparaju, S.~Subramanian, S.~Kalyanakrishnan, and P.~Stone.
\newblock Half field offense: {An} environment for multiagent learning and ad
  hoc teamwork.
\newblock In \emph{AAMAS Adaptive Learning Agents (ALA) Workshop}, 2016.

\bibitem[Parisotto and Salakhutdinov(2017)]{parisotto2017neural}
E.~Parisotto and R.~Salakhutdinov.
\newblock Neural map: Structured memory for deep reinforcement learning.
\newblock arXiv:1702.08360, 2017.

\bibitem[Fulda et~al.(2017)Fulda, Ricks, Murdoch, and Wingate]{fulda2017can}
N.~Fulda, D.~Ricks, B.~Murdoch, and D.~Wingate.
\newblock What can you do with a rock? {Affordance} extraction via word
  embeddings.
\newblock In \emph{Proceedings of the 26th International Joint Conference on
  Artificial Intelligence}, IJCAI'17, pages 1039--1045, 2017.
\newblock \doi{https://doi.org/10.24963/ijcai.2017/144}.

\bibitem[Watkins and Dayan(1992)]{watkins1992q}
J.~C.~H. Watkins and P~Dayan.
\newblock Q-learning.
\newblock \emph{Machine learning}, 8\penalty0 (3-4):\penalty0 279--292, 1992.
\newblock \doi{https://doi.org/10.1007/BF00992698}.

\bibitem[Kostka et~al.(2017)Kostka, Kwiecieli, Kowalski, and
  Rychlikowski]{kostka2017text}
B.~Kostka, J.~Kwiecieli, J.~Kowalski, and P.~Rychlikowski.
\newblock Text-based adventures of the {G}olovin {AI} agent.
\newblock In \emph{2017 IEEE Conference on Computational Intelligence and Games
  (CIG)}, pages 181--188, 2017.
\newblock \doi{https://doi.org/10.1109/CIG.2017.8080433}.

\bibitem[Nachum et~al.(2017)Nachum, Norouzi, and Schuurmans]{UREX}
O~Nachum, M.~Norouzi, and D.~Schuurmans.
\newblock Improving policy gradient by exploring under-appreciated rewards.
\newblock In \emph{5th International Conference on Learning Representations,
  {ICLR} 2017, Toulon, France, April 24-26, 2017, Conference Track
  Proceedings}, 2017.
\newblock arXiv:1611.09321.

\bibitem[Ha(2019)]{Ha_paper}
D.~Ha.
\newblock Reinforcement learning for improving agent design.
\newblock \emph{Artificial Life}, 25\penalty0 (4):\penalty0 352--365, 2019.
\newblock \doi{https://doi.org/10.1162/artl_a_00301}.

\bibitem[Cundy and Filan(2018)]{cundy2018exploring}
C.~Cundy and D.~Filan.
\newblock Exploring hierarchy-aware inverse reinforcement learning.
\newblock arXiv:1807.05037, 2018.

\bibitem[Hessel et~al.(2018)Hessel, Modayil, {Van Hasselt}, Schaul, Ostrovski,
  Dabney, Horgan, Piot, Azar, and Silver]{hessel2018rainbow}
M.~Hessel, J.~Modayil, H.~{Van Hasselt}, T.~Schaul, G.~Ostrovski, W.~Dabney,
  D.~Horgan, B.~Piot, M.~Azar, and D.~Silver.
\newblock Rainbow: {C}ombining improvements in deep reinforcement learning.
\newblock In \emph{Thirty-Second AAAI Conference on Artificial Intelligence},
  2018.

\bibitem[Nachum et~al.(2018)Nachum, Gu, Lee, and Levine]{NIPS2018_7591}
O.~Nachum, S.~Gu, H.~Lee, and S.~Levine.
\newblock Data-efficient hierarchical reinforcement learning.
\newblock In \emph{Advances in Neural Information Processing Systems 31}, pages
  3303--3313. Curran Associates, Inc., 2018.
\newblock URL
  \url{http://papers.nips.cc/paper/7591-data-efficient-hierarchical-reinforcement-learning.pdf}.

\bibitem[Adhikari et~al.(2020)Adhikari, Yuan, C{\^o}t{\'e}, Zelinka, Rondeau,
  Laroche, Poupart, Tang, Trischler, and Hamilton]{adhikari2020learning}
A.~Adhikari, X.~Yuan, M.~A. C{\^o}t{\'e}, M.~Zelinka, M.~A. Rondeau,
  R.~Laroche, P.~Poupart, J.~Tang, A.~Trischler, and W.~L. Hamilton.
\newblock Learning dynamic knowledge graphs to generalize on text-based games.
\newblock arXiv preprint arXiv:2002.09127, 2020.

\bibitem[Kuhlmann and Stone(2003)]{kuhlmann2003progress}
G.~Kuhlmann and P.~Stone.
\newblock Progress in learning 3 vs. 2 {K}eepaway.
\newblock In \emph{IEEE International Conference on Systems, Man and
  Cybernetics}, SMC'03, pages 52--59, 2003.
\newblock \doi{https://doi.org/10.1109/ICSMC.2003.1243791}.

\bibitem[Islam et~al.(2017)Islam, Henderson, Gomrokchi, and
  Precup]{islam2017reproducibility}
R.~Islam, P.~Henderson, M.~Gomrokchi, and D.~Precup.
\newblock Reproducibility of benchmarked deep reinforcement learning tasks for
  continuous control.
\newblock In \emph{ICML Workshop on Reproducibility in Machine Learning},
  ICML'17, 2017.
\newblock arXiv:1708.04133.

\bibitem[Held et~al.(2017)Held, McCarthy, Zhang, Shentu, and
  Abbeel]{held2017probabilistically}
D.~Held, Z.~McCarthy, M.~Zhang, F.~Shentu, and P.~Abbeel.
\newblock Probabilistically safe policy transfer.
\newblock In \emph{IEEE International Conference on Robotics and Automation
  (ICRA)}, pages 5798--5805, 2017.
\newblock \doi{https://doi.org/10.1109/ICRA.2017.7989680}.

\bibitem[Wulfmeier et~al.(2017)Wulfmeier, Posner, and
  Abbeel]{wulfmeier2017mutual}
M.~Wulfmeier, I.~Posner, and P.~Abbeel.
\newblock Mutual alignment transfer learning.
\newblock In \emph{Proceedings of the 1st Annual Conference on Robot Learning},
  2017.

\bibitem[Peng et~al.(2018)Peng, Andrychowicz, Zaremba, and Abbeel]{peng2018sim}
X.~B. Peng, M.~Andrychowicz, W.~Zaremba, and P.~Abbeel.
\newblock Sim-to-real transfer of robotic control with dynamics randomization.
\newblock In \emph{IEEE International Conference on Robotics and Automation
  (ICRA)}, pages 1--8, 2018.
\newblock \doi{https://doi.org/10.1109/ICRA.2018.8460528}.

\bibitem[Oh et~al.(2015)Oh, Guo, Lee, Lewis, and Singh]{NIPS2015_5859}
J.~Oh, X.~Guo, H.~Lee, R.~L. Lewis, and S.~Singh.
\newblock Action-conditional video prediction using deep networks in {A}tari
  games.
\newblock In \emph{Proceedings of the 28th International Conference on Neural
  Information Processing Systems - Volume 2}, NIPS'15, pages 2863--2871, 2015.

\bibitem[Chiappa et~al.(2017)Chiappa, Racaniere, Wierstra, and
  Mohamed]{chiappa2017recurrent}
S.~Chiappa, S.~Racaniere, D.~Wierstra, and S.~Mohamed.
\newblock Recurrent environment simulators.
\newblock arXiv:1704.02254, 2017.

\bibitem[Nagabandi et~al.(2018)Nagabandi, Kahn, Fearing, and
  Levine]{nagabandi2018neural}
A.~Nagabandi, G.~Kahn, R.~S. Fearing, and S.~Levine.
\newblock Neural network dynamics for model-based deep reinforcement learning
  with model-free fine-tuning.
\newblock In \emph{IEEE International Conference on Robotics and Automation
  (ICRA)}, pages 7559--7566, 2018.
\newblock \doi{https://doi.org/10.1109/ICRA.2018.8463189}.

\bibitem[Wang et~al.(2019)Wang, Bao, Clavera, Hoang, Wen, Langlois, Zhang,
  Zhang, Abbeel, and Ba]{wang2019benchmarking}
T.~Wang, X.~Bao, I.~Clavera, J.~Hoang, Y.~Wen, E.~Langlois, S.~Zhang, G.~Zhang,
  P.~Abbeel, and J.~Ba.
\newblock Benchmarking model-based reinforcement learning.
\newblock arXiv:1907.02057, 2019.

\bibitem[Maei and Sutton(2010)]{Maei2010/06}
H.~R. Maei and R.~S. Sutton.
\newblock {GQ(lambda)}: {A} general gradient algorithm for temporal-difference
  prediction learning with eligibility traces.
\newblock In \emph{3rd Conference on Artificial General Intelligence
  (AGI-2010)}. Atlantis Press, 2010.
\newblock ISBN 978-90-78677-36-9.
\newblock \doi{https://doi.org/10.2991/agi.2010.22}.

\bibitem[{van Hasselt} et~al.(2016){van Hasselt}, Guez, and
  Silver]{Hasselt:2016}
H.~{van Hasselt}, A.~Guez, and D.~Silver.
\newblock Deep reinforcement learning with double {Q}-learning.
\newblock In \emph{Proceedings of the Thirtieth AAAI Conference on Artificial
  Intelligence}, AAAI'16, pages 2094--2100, 2016.

\bibitem[Harutyunyan et~al.(2016)Harutyunyan, Bellemare, Stepleton, and
  Munos]{Harutyunyan_2016}
A.~Harutyunyan, M.~G. Bellemare, T.~Stepleton, and R.~Munos.
\newblock Q($\lambda$) with off-policy corrections.
\newblock In \emph{Algorithmic Learning Theory}, pages 305--320, 2016.
\newblock ISBN 978-3-319-46379-7.
\newblock \doi{https://doi.org/10.1007/978-3-319-46379-7_21}.

\bibitem[Quillen et~al.(2018)Quillen, Jang, Nachum, Finn, Ibarz, and
  Levine]{quillen2018deep}
D.~Quillen, E.~Jang, O.~Nachum, C.~Finn, J.~Ibarz, and S.~Levine.
\newblock Deep reinforcement learning for vision-based robotic grasping: A
  simulated comparative evaluation of off-policy methods.
\newblock In \emph{IEEE International Conference on Robotics and Automation
  (ICRA)}, pages 6284--6291, 2018.
\newblock \doi{https://doi.org/10.1109/ICRA.2018.8461039}.

\bibitem[{Bu{\c{s}}oniu} et~al.(2008){Bu{\c{s}}oniu}, {Babu{\v{s}}ka}, and {De
  Schutter}]{4445757}
L.~{Bu{\c{s}}oniu}, R.~{Babu{\v{s}}ka}, and B.~{De Schutter}.
\newblock A comprehensive survey of multiagent reinforcement learning.
\newblock \emph{IEEE Transactions on Systems, Man, and Cybernetics, Part C
  (Applications and Reviews)}, 38\penalty0 (2):\penalty0 156--172, 2008.
\newblock ISSN 1094-6977.
\newblock \doi{https://doi.org/10.1109/TSMCC.2007.913919}.

\bibitem[Zheng et~al.(2018)Zheng, Yang, Cai, Zhou, Zhang, Wang, and
  Yu]{zheng2018magent}
L.~Zheng, J.~Yang, H.~Cai, M.~Zhou, W.~Zhang, J.~Wang, and Y.~Yu.
\newblock {MAgent: A} many-agent reinforcement learning platform for artificial
  collective intelligence.
\newblock In \emph{Thirty-Second AAAI Conference on Artificial Intelligence},
  2018.

\bibitem[Fan et~al.(2018)Fan, Zhu, Zhu, Liu, Zeng, Gupta, Creus-Costa,
  Savarese, and Fei-Fei]{corl2018surreal}
L.~Fan, Y.~Zhu, J.~Zhu, Z.~Liu, O.~Zeng, A.~Gupta, J.~Creus-Costa, S.~Savarese,
  and L.~Fei-Fei.
\newblock {SURREAL}: {O}pen-source reinforcement learning framework and robot
  manipulation benchmark.
\newblock In \emph{Proceedings of The 2nd Conference on Robot Learning},
  volume~87 of \emph{PMLR}, pages 767--782, 2018.

\bibitem[{Lecun} et~al.(1998){Lecun}, {Bottou}, {Bengio}, and
  {Haffner}]{CNNLecun}
Y.~{Lecun}, L.~{Bottou}, Y.~{Bengio}, and P.~{Haffner}.
\newblock Gradient-based learning applied to document recognition.
\newblock \emph{Proceedings of the IEEE}, 86\penalty0 (11):\penalty0
  2278--2324, 1998.
\newblock \doi{https://doi.org/10.1109/5.726791}.

\end{thebibliography}
\end{document}